\documentclass{article}
\pdfoutput=1

\usepackage[preprint]{neurips_2023}

\usepackage[utf8]{inputenc} % allow utf-8 input
\usepackage[T1]{fontenc}    % use 8-bit T1 fonts
\usepackage{hyperref}       % hyperlinks
\usepackage{url}            % simple URL typesetting
\usepackage{booktabs}       % professional-quality tables
\usepackage{amsfonts}       % blackboard math symbols
\usepackage{nicefrac}       % compact symbols for 1/2, etc.
\usepackage{microtype}      % microtypography
\usepackage{caption}
\usepackage{booktabs} % for professional tables
\usepackage{multirow}
\usepackage{mathtools}
\usepackage{bm,bbm}
\usepackage{hyperref}
\hypersetup{colorlinks=true,breaklinks=true,citecolor={blue!80!black}}
\usepackage{algorithm}
\usepackage{algorithmic}
\usepackage{setspace}
\usepackage{listings}
\usepackage{graphicx}
\usepackage{subcaption}
\usepackage{enumitem}
\usepackage[table,xcdraw]{xcolor}

\newcommand{\inputseq}{\mathbf{X}}
\newcommand{\timevector}{\mathbf{x}}
\newcommand{\outputseq}{\mathbf{\hat{X}}}
\newcommand{\timeprediction}{\mathbf{\hat{x}}}
\newcommand{\encoderout}{\mathbf{Z}}
\newcommand{\STFTout}{\Tilde{\mathbf{X}}}
\newcommand{\TFBout}{\mathbf{Q}}

\newcommand{\kernelmatrix}{\mathbf{W}}
\newcommand{\hidden}{\mathbf{H}}
\newcommand{\predlen}{T}
\newcommand{\inputlen}{L}
\newcommand{\channels}{D}

\newcommand{\complex}{C}
\newcommand{\nfft}{M}
\newcommand{\timeframe}{N}
\newcommand{\windowsize}{S}
\newcommand{\frameshift}{l}
\newcommand{\gate}{\mathbf{g}}
\newcommand{\gatematrix}{\mathbf{G}}

\title{TFDNet: Time-Frequency Enhanced Decomposed Network for Long-term Time Series Forecasting}

\author{%
  Yuxiao~Luo \\
  The Chinese University of Hong Kong, Shenzhen \\
  Shenzhen, China \\
  \texttt{yuxiaoluo@link.cuhk.edu.cn} \\
  \And
  Ziyu Lyu \\
  Shenzhen Institutes of Advanced Technology, Chinese Academy of Sciences \\
  Shenzhen, China \\
  \texttt{zy.lv@siat.ac.cn} \\
  \AND
  Xingyu Huang \\
  University of Science and Technology of China \\
  Shenzhen, China \\
  \texttt{huangxyu@mail.ustc.edu.cn} \\
}

\begin{document}

\maketitle

\begin{abstract}
Long-term time series forecasting is a vital task and has a wide range of real applications. Recent methods focus on capturing the underlying patterns from one single domain (e.g. the time domain or the frequency domain), and have not taken a holistic view to process long-term time series from the time-frequency domains. In this paper, we propose a \textbf{T}ime-\textbf{F}requency \textbf{E}nhanced \textbf{D}ecomposed \textbf{N}etwork (\textbf{TFDNet}) to capture both the long-term underlying patterns and temporal periodicity from the time-frequency domain. In TFDNet, we devise a multi-scale time-frequency enhanced encoder backbone and develop two separate trend and seasonal time-frequency blocks to capture the distinct patterns within the decomposed trend and seasonal components in multi-resolutions. Diverse kernel learning strategies of the kernel operations in time-frequency blocks have been explored, by investigating and incorporating the potential different channel-wise correlation patterns of multivariate time series.
Experimental evaluation of eight datasets from five benchmark domains demonstrated that TFDNet is superior to state-of-the-art approaches in both effectiveness and efficiency.
 %Long-term time series forecasting is a vital task and has a wide range of real applications. Recent methods focus on capturing the underlying patterns from one single domain (e.g. the time domain or the frequency domain), and have not taken a holistic view to process long-term time series in the time-frequency domains. In this paper, we propose a \textbf{T}ime-\textbf{F}requency \textbf{E}hanced \textbf{D}ecomposed \textbf{N}etwork or \textbf{TFDNet} to extract the historical information from the time-frequency domain, and devise multi-scale time frequency enhanced encoders for the decomposed trend and seasonal components. Two separate trend and seasonal time-frequency blocks are devised to capture the distinct patterns within the decomposed trend and seasonal components. Different learning strategies are developed in the trend and seasonal time-frequency blocks, by investigating and integrating the channel-wise correlation effects of multivariate time series. Experimental evaluations on eight datasets from five benchmarks demonstrate that TFDNet is superior to state-of-the-art approaches in both effectiveness and efficiency.
\end{abstract}
\section{Introduction}
Time series forecasting is a long-standing task, and has been widely used in various application domains, e.g. energy, weather, transportation, and economics \cite{rajagukguk2020review,angryk2020multivariate,chen2019review,sezer2020financial}. Especially, long-term time series forecasting is a challenging problem and has attracted more attention \cite{wu2021autoformer,zhou2022fedformer}.  In recent decades, deep learning methods have made remarkable innovations in time series forecasting,  from recurrent neural network (RNN) based methods \cite{zhang1998time,graves2012long,chung2014empirical} and temporal convolution networks \cite{sen2019think,borovykh2017conditional,bai2018empirical} to the recent transformer-based models \cite{zhou2021informer,wu2021autoformer,zhou2022fedformer,nie2022time}. However, the computation complexity and memory requirements of transformer-based methods make it difficult to perform long sequence modeling. Recently, simple Linear models like DLiner \cite{zeng2022transformers} have been devised and achieved competitive performance with transformer-based solutions.

Despite the successful progress of diverse learning models, some issues have not been fully answered. First, the \textbf{time-frequency} information of long-term time series has not been fully studied. Most of the existing solutions focused on processing information from a single domain, e.g. the time domain or the frequency domain. For example, Autoformer \cite{wu2021autoformer} and PatchTST \cite{nie2022time} only considered temporal periodicity and semantic learning via empowering transformer structures. FiLM \cite{zhou2022film} concentrated on the frequency domain. Both time and frequency information are important for time series analysis and forecasting. The time domain includes temporal correlation and temporal periodicity, and the frequency domain can capture the global properties and the underlying change patterns of time series.  
%Second, the \textbf{channel-wise effects} of multivariate time series has not been well discussed, although some studies \cite{nie2022time,zeng2022transformers} have shown that adopting the channel-independence (CI) strategy can have better performance than utilizing the channel-mixing (CM) strategy to allow interactions among channels. The two strategies have different assumptions that strong correlations (CM) or weak correlations (CI) among channels of multivariate time series. In fact, the channel-wise correlations depend on the inherent characteristics of the time series data in different domains. But previous studies have not analyzed and investigated the channel-wise correlations when designing the model structure.
Second, although some studies \cite{nie2022time,zeng2022transformers} have shown that adopting the channel-independence (CI) strategy can have better performance than utilizing the channel-mixing (CM), the \textbf{channel-wise correlation effects} has not been well discussed in CI strategy. Different from that the CM strategy embeds the whole channels into high-dimensional vectors to represent the entire information of all channels, while the input of CI strategy only considers the information from the single channel. In this case, channel-wise correlation is necessary to be considered when designing the model structure. In fact, the channel-wise correlations depend on the inherent characteristics of the time series data in different domains. However previous studies utilized shared parameters for all channels without considering their channel-wise correlations.

\begin{figure}[t]
  \centering
  \includegraphics[width=0.8\textwidth]{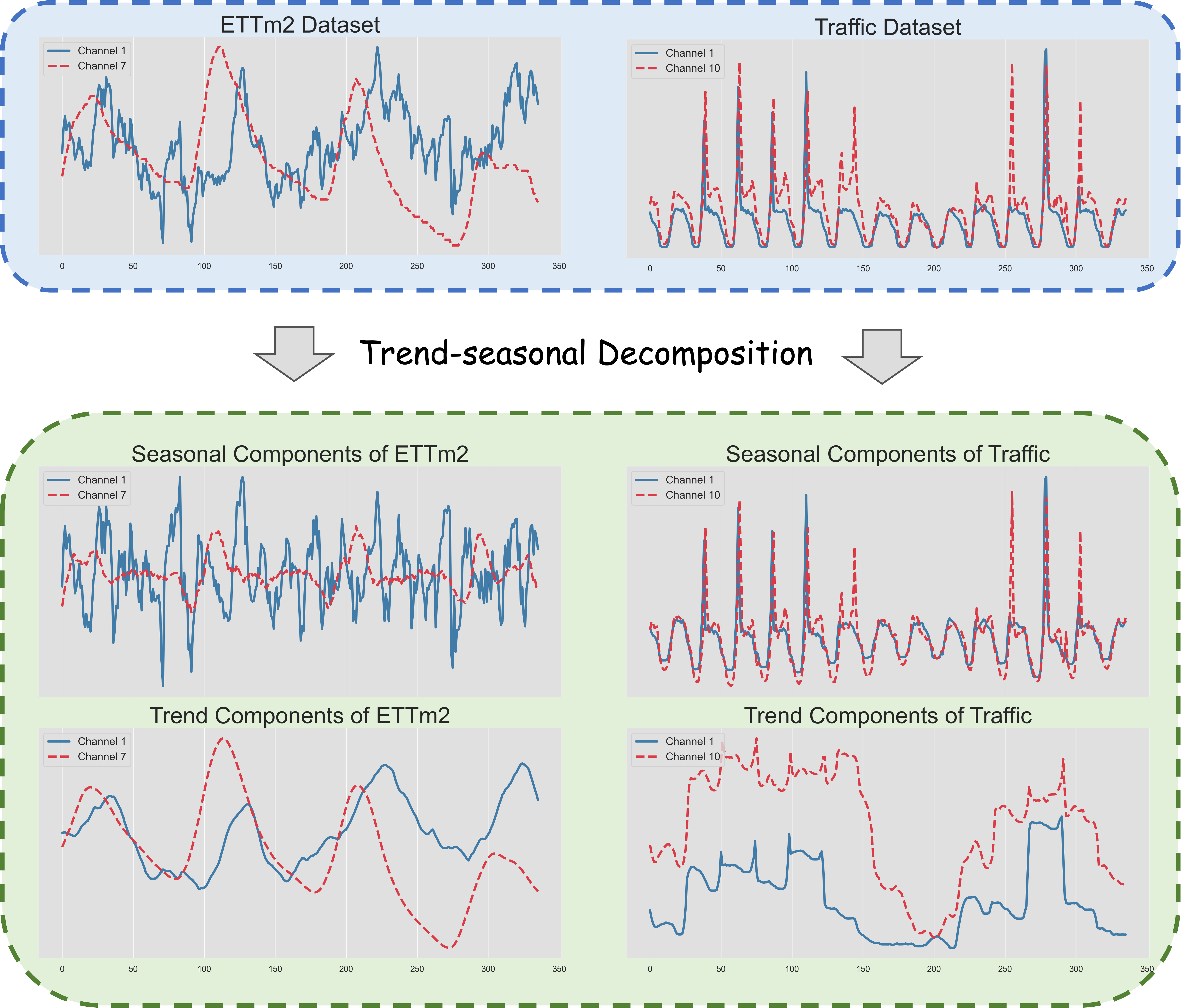}
  % \caption{Overall structure of TFDNet. The TFDNet is comprised of two branches, the seasonal branch and the trend branch.$K$ in the Mixture of Experts for Multiple Short-Time Fourier Transform Windows (MSTFT-MOE) donates the STFT window sizes. Seasonal Encoder and Trend Encoder share a common backbone with Time-Frequency Block (TFB) and Frequency Feed Forward Network (FFFN) as depicted in the right figure. Seasonal Encoder employs Seasonal TFB*, which has two subversions: Seasonal TFB Individual Kernel (STFB-IK) and Seasonal TFB Multiple Kernel (STFB-MK). The Trend Encoder utilizes the Trend TFB (TTFB) to process the trend component. The figure on the left illustrates the domain transformation process employed by TFDNet.}
  % This process involves the conversion of input signals from the time domain to the time-frequency domain, followed by the reconstruction of the output signals back to the time domain.
  \caption{The top two subplots showcase the raw signals of ETTm2 and Traffic from two different channels. The lower four subplots depict the decomposed seasonal and trend components of the two datasets.}
  \label{fig:channel_corr}
\end{figure}
To address the above challenges, we propose a \textbf{T}ime-\textbf{F}requency \textbf{E}nhanced \textbf{D}ecomposed \textbf{N}etwork (\textbf{TFDNet}) for long-term time series forecasting.  \textbf{First}, we take a holistic view to process long-term time series in the time-frequency domain, by leveraging the classical signal processing method Short-time Fourier transform (STFT) \cite{boashash2015time} to transform long-term time series into time-frequency matrix.  \textbf{Second}, we analyze the channel-wise correlations for the whole multivariate time series in diverse domains and further investigate channel-wise correlations within the trend component and the seasonal component decomposed with the trend-seasonal decomposition block. 
%\ziyu{The seasonal components typically show more complex patterns and frustration \cite{wang2023micn,zhang2022detrend}, so we focus on these parts. As shown in Table~\ref{tab:MACC}, we observe some time series data demonstrate lower channel-wise correlations in the seasonal component compared with the original data, e.g. Weather. For these datasets, it is possible to consider that the seasonal part exhibits a greater amount of distinctive information across channels. While other datasets like Traffic exhibit stronger correlations in season, thereby these parts augment their mutual information. We consider the channel-wise effects in our design, especially for the seasonal components.} 
Figure~\ref{fig:channel_corr} illustrates various channel-wise correlations among seasonal components and trend components using the ETTm2 and Traffic datasets as an example. Regarding the trend components, there are clear relationships between the channels in these two datasets. In comparison, whereas the seasonal components of ETTm2 are not significantly correlative, the seasonal components of Traffic exhibit a substantial association.
% As shown in Figure~\ref{fig:channel_corr},  We observe that the channel-wise correlations of the whole multivariate time series are actually different in diverse domains, e.g. high correlations in Traffic dataset. There might be different patterns between the trend component and the seasonal component e.g. higher channel-wise correlations in the seasonal component than the trend component for Traffic dataset and the contrary pattern for Weather dataset (lower correlations in the seasonal component). 
We consider the channel-wise effects in our design of processing the trend and seasonal components. By integrating the two ideas, we devise multi-scale time-frequency enhanced encoders in \textbf{TFDNet} for the trend and seasonal components. The multi-scale windowing mechanism is devised to capture the time-frequency information in diverse resolutions. Two separate time-frequency blocks (TFBs) are respectively developed for the trend component and the seasonal component due to the distinct underlying patterns within them. In addition, we integrate the observation of different channel-wise correlation patterns into kernel operations of time-frequency blocks and introduce one individual kernel strategy and a multiple kernel sharing strategy to deal with various channel-wise correlation patterns in the seasonal component. Finally, we devise a mixture loss to enable robust forecasting, by combining the L1 loss and L2 loss. The main contributions are listed as follows:
\begin{itemize} 
    \item A Time-Frequency Enhanced Decomposed Network is proposed for robust and effective long-term time series forecasting, by taking a holistic view to process long-term time series from the time-frequency domain.
    \item We investigate the channel-wise correlation effects of multivariate time series forecasting, and integrate the channel-wise correlation effects into the design of separate time-frequency blocks for the trend and seasonal components.
    \item We conduct extensive experiments on 8 datasets in various domains, e.g. traffic, weather, electricity, etc. The experimental results demonstrated that our method is superior to state-of-the-art methods with high efficiency. 
\end{itemize}

\section{Related work}
Time series forecasting is an important task and has a wide range of applications, e.g. traffic flow prediction, weather prediction, and energy estimation \cite{chen2019review,rajagukguk2020review}. Many methods have been proposed for time series forecasting, from the traditional statistical methods (e.g. ARIMA model \cite{ho1998use}) to machine learning methods. Especially several branches of deep learning methods have been devised for long-term time series forecasting, including recurrent neural networks \cite{zhang1998time,graves2012long,chung2014empirical}, temporal convolution networks \cite{sen2019think,borovykh2017conditional,bai2018empirical}, and the recent transformer-based solutions \cite{vaswani2017attention,wu2021autoformer,zhou2021informer}. We describe existing studies based on the utilized information domain (e.g. time domain and frequency domain) and the channel strategies of multivariate time series prediction in the following paragraphs.

\paragraph{Representation in time domain}
The temporal correlations and temporal periodicity play a vital role in time series analysis and forecasting. Tansformer-based solutions have been proposed to capture the long-range dependencies within the time domain. However, transformer-based solutions have some limitations, e.g. the model complexity and large memory requirements. 
Informer \cite{zhou2021informer} incorporated KL-divergence-based ProbSparse attention to reduce the model complexity. Autoformer \cite{wu2021autoformer} renovated the transformer into a deep decomposition structure and designed the auto-correlation mechanism to capture temporal periodicity at the sub-series level. Non-stationary Transformer \cite{liu2022non} proposed series stationarization and de-stationary Attention for over-stationarization. TimesNet \cite{wu2022timesnet} analyzed temporal variations from the 2D space within multiple periods.

\paragraph{Representation in frequency domain}
The underlying patterns in the frequency domain are important for time series forecasting \cite{sun2022fredo}. Therefore, several methods have leveraged frequency enhanced structure for time series forecasting. For example, FEDformer \cite{zhou2022fedformer} employed a Fourier-enhanced structure to obtain frequency domain mapping. ETSformer \cite{woo2022etsformer} utilized exponential smoothing attention and frequency attention to replace the self-attention mechanism in Transformer. FiLM \cite{zhou2022film} employed Legendre Polynomials projections \cite{voelker2019legendre} to approximate historical information with Fourier projection for eliminating any extraneous noise.

\textbf{Channel-independence strategy} Channel-independence strategy is characterized in PatchTST \cite{nie2022time} that each input tokens only contain information from a single channel univariate time series. PatchTST \cite{nie2022time} divided time series into subseries-level patches, served as input tokens to Transformer. DLinear \cite{zeng2022transformers} challenged the effectiveness of transformer-based structures, and introduced a set of simple linear models layer for each channel. 
Channel-independence models can extract more historical information from a longer length of the look-back window, which increases the capacity of the model \cite{han2023capacity}. 

% Our TFDNet uses a channel-independence strategy. However, the difference between the before works is that TFDNet considers channel correlations. As for seasonal components, TFDNet-IK employs distinct parameters for each channel of the time series that exhibit low channel correlation, while TFDNet-MK utilizes common parameters for channels that demonstrate high channel correlation.

% \input{section/3_background}

\section{Time-frequency enhanced decomposed
network}
\label{sec:method}
\subsection{TFDNet framework}
\label{sec:framework}
%\ziyu{symbols follow rules: We use lower-case fonts for scalars, bold lower-case fonts for vectors and bold upper-case fonts for matrices. For example, $p$ is a scalar, $\mathbf{p}$ is a vector and $\mathbf{P}$ is a matrix. matrix product uses $\cdot$, element-wise uses the Hadamard product ($\odot$) }
\paragraph{Preliminary}
Long-term time series forecasting is a sequence-to-sequence problem. Given the historical time series data $\inputseq=[\timevector_1, \timevector_2,\dots,\timevector_L]$ with $L$ length look-back window and each time vector $\timevector_t$ at the $t$ time step represents the multivariate time point with $D$ dimension (the number of channels in the multivariate time series), the problem is to predict the future time series $\outputseq=[\timeprediction_{L+1},\dots,\timeprediction_{L+T}]$, $\outputseq \in \mathbb{R}^{T \times D}$ at $T$ future time steps ($T>1$).

\begin{figure*}[t]
  \centering
  \includegraphics[width=0.95\textwidth]{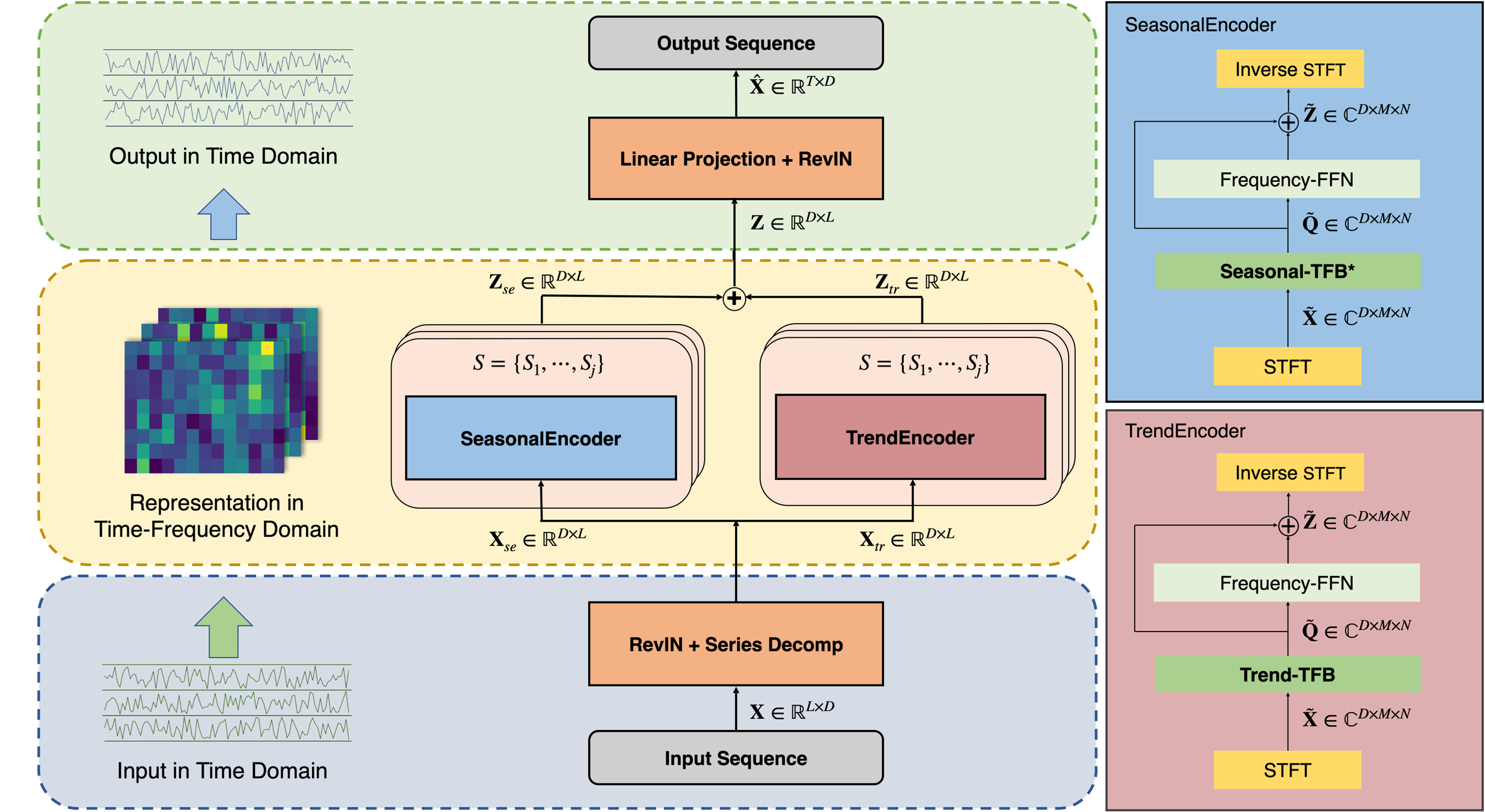}
  % \caption{Overall structure of TFDNet. The TFDNet is comprised of two branches, the seasonal branch and the trend branch.$K$ in the Mixture of Experts for Multiple Short-Time Fourier Transform Windows (MSTFT-MOE) donates the STFT window sizes. Seasonal Encoder and Trend Encoder share a common backbone with Time-Frequency Block (TFB) and Frequency Feed Forward Network (FFFN) as depicted in the right figure. Seasonal Encoder employs Seasonal TFB*, which has two subversions: Seasonal TFB Individual Kernel (STFB-IK) and Seasonal TFB Multiple Kernel (STFB-MK). The Trend Encoder utilizes the Trend TFB (TTFB) to process the trend component. The figure on the left illustrates the domain transformation process employed by TFDNet.}
  % This process involves the conversion of input signals from the time domain to the time-frequency domain, followed by the reconstruction of the output signals back to the time domain.
  \caption{TFDNet architecture. The TFDNet is comprised of two branches, the seasonal branch, and the trend branch. The two branches share a common backbone with Time-Frequency Block and Frequency Feed Forward Network (Frequency-FFN) through multi-scale time-frequency enhanced encoders of $\windowsize$ various window sizes. }
  \label{fig:overall_structure}
\end{figure*}

\paragraph{TFDNet structure}
The proposed TFDNet structure is shown in Figure~\ref{fig:overall_structure} and TFDNet has three phases. In the first phase, we pre-process the input time series for further learning and the process of the first phase is defined as Equation~\ref{eqn:first}.
We adopt of instance normalization \cite{kim2021reversible,liu2022non} 
the historical time series $\inputseq$ to mitigate the distribution shift effects between the training and test data $\inputseq =\text{RevIN}(\inputseq)^\mathsf{T}$, in which each univariate time series $\timevector^{(i)}$ is normalized with zero mean and standard deviation. 
%Transposing $\inputseq$ makes it easier to analyze the channel independently.
After instance normalization, the series decomposition block in Autoformer \cite{wu2021autoformer} is exploited to decompose time series into the seasonal and trend components for capturing the different underlying patterns within the seasonal and trend parts.
%In order to extract the historical information from 1D time series data within the 2D time-frequency representation and consideration the different properties between the seasonal component and the trend component, A newly proposed model, named TFDNet, is presented in this study. By utilizing Multi-Scale STFT (MSTFT), we aim to address the limitations of the traditional approach, which uses a single and fixed window for Short-Time Fourier Transform (STFT).
%As depicted in Figure \ref{fig:overall_strcuture}, the input multivariate time series denoted $\mathcal{X}\in{\mathbb{R}^{L\times{D}}}$. Following instance normalization \cite{kim2021reversible,liu2022non}, we transform the dimension of $\mathcal{X}$ from $L\times{D}$ into $D\times{L}$ making it easier to analyze the channel independently. There is a series decomposition block \cite{wu2021autoformer} to decompose the series into seasonal and trend parts by moving average with with the padding operation. It can be formalized as 
\begin{equation}
\begin{split}
\label{eqn:first}
    %\inputseq &=\text{RevIN}(\inputseq)^\mathsf{T},\\
    \inputseq_{tr}&=\text{AvgPool}(\text{Padding}(\inputseq)),\\
    \inputseq_{se}&=\inputseq-\inputseq_{tr}, \\
\end{split}
\end{equation}
where $\inputseq_{se}$ and $\inputseq_{tr}$ respectively denote the decomposed seasonal component and trend component. $\text{AvgPool}$ is the moving average operation to smooth out periodic hidden variables with the padding operation to make the series length unchanged.
%Following the decomposition of the input signal into two components, we employ Trend Encoder and Seasonal Encoder to process them individually, owing to the observed dissimilarity in patterns between the two components. The process is:

In the second phase, we devise multi-scale time-frequency encoders to separately capture the underlying time and frequency patterns within the seasonal and trend components.
The general process of multi-scale time-frequency encoders is defined in Equation~\ref{eqn:mtf-encoder}, and more design details are illustrated in Section~\ref{sec:mtf-encoder}.
\begin{equation}
\begin{split}
    \mathcal{\encoderout}_{se}=\text{Linear}(&\textbf{SeasonEncoder}(\inputseq_{se}, S_1), \dots,\textbf{SeasonEncoder}(\inputseq_{se}, S_s)),\\
    \mathcal{\encoderout}_{tr}=\text{Linear}(&\textbf{TrendEncoder}(\inputseq_{tr}, S_1),\dots,\textbf{TrendEncoder}(\inputseq_{tr}, S_s)),\\
    %\encoderout&=\encoderout_{se}+\encoderout_{tr},\\
\label{eqn:mtf-encoder}
\end{split}
\end{equation}
in which \textbf{TrendEncoder} and \textbf{SeasonalEncoder} are the two devised encoders, respectively represented with blue and pink boxes in Figure~\ref{fig:overall_structure}. $S_1,\dots, S_s$ represent the utilized different sizes of window lengths. 
One linear layer is adopted to fuse the multi-scale seasonal representations from different scales of seasonal encoders and outputs the fused seasonal representation $\encoderout_{se}$. Similarly, we obtain the fused trend representation $\encoderout_{tr}$. Finally, we sum up the seasonal representation and the trend representation as $\encoderout=\encoderout_{se}+\encoderout_{tr}$ for the further prediction phase.
%where$\mathcal{\encoderout}_{se}\in{\mathbb{\real}^{\channels\times{\inputlen}}}$ and $\mathcal{\encoderout}_{tr}\in{\mathbb{\real}^{\channels\times{\inputlen}}}$ correspond to the results obtained from MSTFT of Seasonal Encoder and MSTFT of Trend Encoder respectively. To reconstruct the series, we sum them together directly.

The final phase is to predict the future time series with $T$ time steps, based on the fused encoder representation. The prediction process is defined as follows: $\outputseq=\text{RevIN}(\text{Linear}(\encoderout))$.
A linear projection and instance denormalization \cite{kim2021reversible} are used to predict the future time series from the fused encoder representation $\encoderout$.

\subsection{Multi-scale Time-frequency Enhanced Encoders}
\label{sec:mtf-encoder}
We adopt the multi-scale strategy to adjust the sliding window size of Short-time Fourier Transform when analyzing the time-frequency domain. As mentioned in Section~\ref{sec:framework}, we set a different sliding window size $S_e$ for each time-frequency $\text{encoder}_{e}$ and learn the time-frequency representation in multi-scale resolution. For both the seasonal encoder and the trend encoders, we have set the multi-scale sliding window sizes with $\windowsize=\{\windowsize_1,\cdots,\windowsize_s\}$. With the given window size $S_e$, the $e$-th trend encoder and the $e$-th seasonal encoder share the same backbone structure except for the time-frequency block (TFB). For simplicity, we omit the encoder index with different scales ($S$) in the following paragraphs. 
In the following subsections, we first introduce the devised encoder structure and then demonstrate the different time-frequency blocks for the trend component (Trend-TFB) and the seasonal component (Seasonal-TFB).
\subsubsection{Time-frequency enhanced encoder structure}
\label{sec:encoder}
The decomposed representation $\encoderout_{tr}$ or $\encoderout_{se}$ is the input into the time-frequency enhanced encoder. As the trend encoder and the seasonal encoder share the same encoder structure, we omit the subscripts ($tr$ and $se$) in Section~\ref{sec:encoder}. The Time-frequency enhanced encoder structure has four processing parts, including Short-Time Fourier Transform (STFT), Time-frequency Block, Frequency Feed Forward Network (frequency-FFN), and Inverse Short-Time Fourier Transform (Inverse-STFT). The encoder process is defined as in Equation~\ref{eqn:encoder}
%where there are two blocks called Time-frequency Block (TFB) and Frequency Feed Forward Network (FFFN). The TFB is used to extract the time-frequency information from STFT, and FFFN is one fully connected layer with the activation function Tanh \cite{fan2000extended} processing the frequency information inside of one window. It is formalized as
\begin{equation}
\begin{split}
\STFTout &=\text{STFT}(\inputseq),\\
\TFBout &=\text{TFB}(\STFTout),\\
\tilde{\encoderout}&=\TFBout+\text{Frequency-FFN}(\TFBout),\\
\encoderout&=\text{STFT}^{-1}(\tilde{\encoderout}).\\
\end{split}
\label{eqn:encoder}
\end{equation}

\paragraph{Short-time Fourier transform:} We leverage a classical method Short-time Fourier transform (STFT) to analyze the frequency content of a non-stationary signal and observe its temporal evolution \cite{parchami2016recent,griffin1984signal}. STFT enables the transformation of time series from the time domain to the time-frequency domain, which segments the input time series into overlapping frames and performs Discrete Fourier Transform (DFT) on each frame. The STFT process is defined as:
\begin{equation}
\tilde{\inputseq}= \sum^{\windowsize-1}_{m=0} \text{Window}[m]\inputseq[m+n{l}]e^{-j\frac{2\pi{m}\omega}{\windowsize}},
\end{equation}
% $\tilde{X}[\omega,\tau]= \sum^{\windowsize-1}_{m=0} w[m]x[m+n{l}]e^{-j\frac{2\pi{n}\omega}{\windowsize}}$. \ziyu{edit}.
where the window function is treated as if having $1$ everywhere in the window. The output matrix from the STFT operation $\STFTout\in \mathbb{\complex}^{\channels\times{\nfft}\times{\timeframe}}$ denotes the outputted time-frequency matrix from $\text{STFT}(\inputseq)$. 
$\nfft=\frac{S}{2}+1$ denotes the size of the components of the Fourier transform according to the conjugate symmetry \cite{dubois1978convolution}. $\timeframe$ represents the number of time frames, and $\timeframe=\frac{L}{\frameshift}+1$ by padding the window length S. $\frameshift$ is the stride of the moving window when performing STFT.
%With padding the window length $\windowsize$, $N$ can be calculated as $\inputlen/\frameshift+1$. It is noticeable that the dimension of $\tilde{\mathcal{\tfbout}}$ is the same as $\tilde{\mathcal{\inputseq}}$. As illustrated in Figure \ref{fig:overall_strcuture}, the two modules exhibit distinct configurations for TFBs, and more details in the next section. 

\paragraph{Time-frequency block:}
The Time-frequency Block (TFB) is utilized within encoders to capture the underlying patterns within the time-frequency domain.
We devise the kernel operation on identical frequency bins across various time frames. 
For each channel ($i$-th channel) of the STFT output $\STFTout^{(i)}\in{\mathbb{\complex}^{\nfft\times{\timeframe}}}$, the kernel operation is generally defined as follows:
\begin{equation}
\begin{split}
    \textbf{Kernel}(\STFTout^{(i)}, \kernelmatrix)=\STFTout^{(i)}_{m}\cdot \kernelmatrix_m,
\end{split}
\label{eqn:kernel}
\end{equation}
in which $\kernelmatrix \in\mathbb{\complex}^{\nfft\times\timeframe\times\timeframe}$ is the kernel weight matrix. $\STFTout^{(i)}_{m} \in\mathbb{\complex}^{\timeframe}$ donates the $m$-th frequency bin and $\kernelmatrix_m \in\mathbb{\complex}^{\timeframe\times\timeframe}$ donates corresponding weights for the $i$-th channel.
We devise different strategies to learn the kernel weight matrix $\kernelmatrix$ for the trend and seasonal components, according to the inherent properties of different components. 
More details about the design of TFB for the trend and seasonal components are demonstrated in Section~\ref{sec:TFB}.

\paragraph{Frequency Feed Forward Network:} Frequency-FFN is one fully connected layer with the activation function Tanh \cite{fan2000extended} to accumulate the time-frequency information along the \textbf{frequency} side within one time step. We combine the processed $\text{Frequency-FFN}(\TFBout)$ and $\TFBout$ to produce the time-frequency encoding $\tilde{\encoderout}$.

\paragraph{Inverse Short-time Fourier transform:} $\inputseq$ can be reconstructed from $\tilde{\inputseq}$, by taking the inverse Discrete Fourier Transform (DFT) of each DFT vector and overlap adding the inverted signals \cite{parchami2016recent}.

\subsubsection{Time-frequency block}
\label{sec:TFB}
\begin{figure*}[t]
  \centering
  \includegraphics[width=0.9\textwidth]{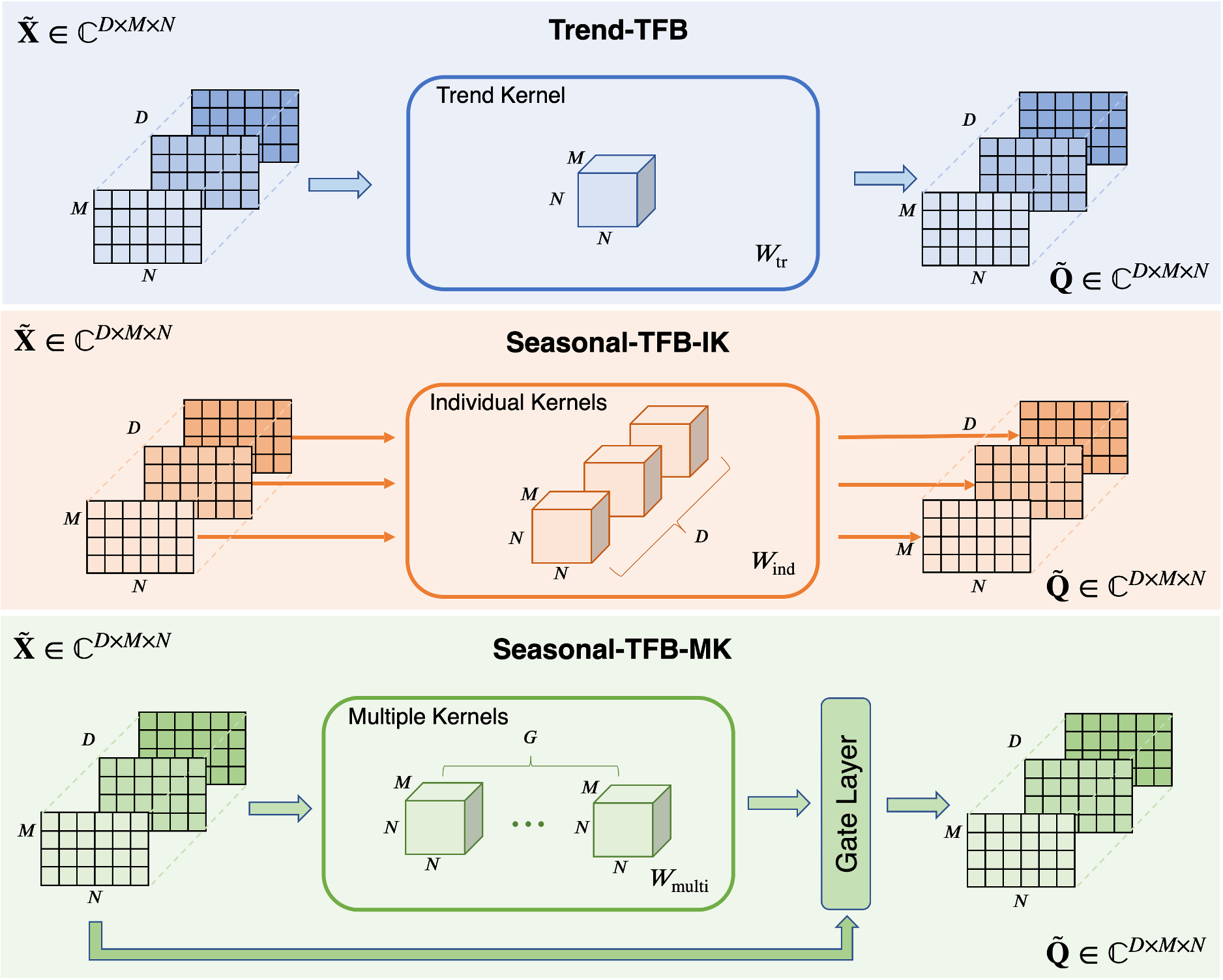}
  \caption{Workflow of Time-frequency Block. Trend-TFB is used in TrendEncoder (Top). Seasonal-TFB with individual kernels (Seasonal-TFB-IK) and seasonal-TFB with multiple shared kernels (Seasonal-TFB-MK) are used in SeasonalEncoder.}
  \label{fig:TFB_structure}
\end{figure*}

Two separate time-frequency blocks are devised for the trend component and the seasonal component, respectively Trend-TFB and Seasonal-TFB. The design of time-frequency blocks is shown in Figure~\ref{fig:TFB_structure}.

\paragraph{Trend time-frequency block:}
Previous studies have found that the long-term underlying pattern of the trend component is relatively simple, compared to the seasonal component \cite{wang2023micn}. Therefore, we devise a simple and effective Trend-TFB to process the trend component, as shown in the top figure of Figure~\ref{fig:TFB_structure}. We utilize one single shared kernel to process the trend pattern across multiple channels. We obtain $\TFBout_{tr}^{(i)}=\textbf{Kernel}(\STFTout^{(i)}, \kernelmatrix_{tr})$ by learning the $\kernelmatrix^{(i)}_{tr}$ for each univariate time sequence at the $i$-th channel with the single shared kernel. The final output of Trend-TFB is $\TFBout_{tr} \in{\mathbb{\complex}^{\channels\times{\nfft}\times{\timeframe}}}$.

\paragraph{Seasonal time-frequency block:}
From the analysis studies in Figure~\ref{fig:channel_corr}, we observe that datasets in different domains have different seasonal characteristics, e.g. high seasonal correlation in Traffic, and a relatively lower correlation in ETTm2. Therefore, we design two versions to address the potential effects due to different seasonal correlation patterns among channels, respectively \textbf{seasonal-TFB with individual kernels (Seasonal-TFB-IK)} and \textbf{seasonal-TFB with multiple shared kernels (Seasonal-TFB-MK}). The experimental results validate the essence of the specific design (individual or shared) for datasets in different domains. Different designs of the individual version (middle) and the shared version (bottom) are shown in Figure~\ref{fig:TFB_structure}.

Seasonal-TFB-IK adopts the individual kernel strategy without sharing kernel weights across different channels. However, it is intractable when the channel is huge. Therefore, we leverage the low-rank approximation techniques \cite{yuan2009sparse,hyndman2018forecasting} to make Seasonal-TFB-IK scalable.
For a large number of channels, the kernel matrix $\kernelmatrix_{ind} \in{\mathbb{\complex}^{\channels\times{\nfft}\times{\timeframe}\times{\timeframe}}}$ is decomposed with two smaller matrices $\kernelmatrix_1\in{\mathbb{\complex}^{\channels\times{I}}}$ and $\kernelmatrix_2\in{\mathbb{\complex}^{I\times{\nfft}\times{\timeframe}\times{\timeframe}}}$ ($I\leq{\channels})$, and $\kernelmatrix_{ind}=\kernelmatrix_1 \cdot \kernelmatrix_2$. The two smaller kernel matrices will be learned, and we have $\TFBout_{se}^{(i)}=\textbf{Kernel}(\STFTout_{se}^{(i)},\kernelmatrix_{ind}^{(i)})$ for the $i$-th channel of the multivariate time series from Seasonal-TFB-IK.

For the data domain with high seasonal correlation, Seasonal-TFB-MK is introduced, in which we devise a multiple kernel sharing strategy and utilize one gate layer to combine the multiple kernels. Each channel is processed by multiple kernels, which can extract multiple underlying patterns across channels within the seasonal component. 
%STFB has two sub-versions of individual kernel parameters for each channel and shared multiple kernel parameters with gates, called Seasonal TFB Individual Kernel (STFB-IK) and Seasonal TFB Multiple Kernel (STFB-MK). These settings are because some time series have obvious low correlations across channels in the seasonal part, in this case, individual parameters are more suitable. By comparison, STFB-MK has better performance when time series showthe high correlation in the seasonal parts of different channels with complicated seasonal fluctuation.
When having $k$ shared kernels, we learn $\{\kernelmatrix_{multi}^1,\cdots,\kernelmatrix_{multi}^k\}$ through the kernel operation for the multivariate time series. For the i-th channel time series, the output from one kernel is $\hidden_k^{(i)}=\textbf{Kernel}(\STFTout^{(i)}, \kernelmatrix^k_{multi})$, and we have multiple kernel representations $\{\hidden_1^{(i)}, \cdots,\hidden_k^{(i)}\}$. Next, the gate vector $\gate$ produced by $\gatematrix^k\in{\mathbb{\complex}^{\nfft\times{\timeframe}}}$ in the gate layer is utilized to fuse the multiple kernel representations. The $k$-th element in the gate vector is defined as
$\gate_k=\text{Sigmod}(|\sum\gatematrix^k \odot \STFTout_{se}^{(i)}|)$.
Finally, $\TFBout_{se}^{(i)}=\sum_k\gate \odot [\hidden_1^{(i)};\hidden_2^{(i)};\cdots \hidden_k^{(i)}]$ for  the $i$-th channel of the multivariate time series from Seasonal-TFB-MK.
% $\gatematrix\in{\mathbb{\complex}^{\nfft\times{\timeframe}}}$.

% The gate layer is to learn the kernel weights to fuse multiple kernels for each channel.
\subsection{Mixture loss}
\label{loss}
The squared loss (L2) has been widely used for time series prediction \cite{wu2021autoformer,zhou2022fedformer,nie2022time}. However, the squared loss is sensitive to occasional outliers. In order to improve the robustness of the model, we combine the squared loss and the absolute loss (L1) as the mixture loss, as defined in Equation~\ref{eqn:loss}.
% High incidence of noise in real-time sequences together with higher requirements for inference robustness.
% It has been observed that the Mean Squared Error (MSE) loss function is commonly utilized in many time series forecasting models to measure the dissimilarity between the predicted and actual values \cite{wu2021autoformer,zhou2022fedformer,nie2022time}. However, MSE Loss is sensitive to outliers which makes it is easier to affected by noises. Noise is abundant in real-time series, in this case, sensitivity to noise reduces the robustness of the model. By comparison, the Mean Absolute Error (MAE) loss has better robustness in the time series forecasting task resulting in more accurate predictions \cite{han2023capacity}. However, it reduces the accuracy in periodic time series like Traffic and Electricity datasets \cite{han2023capacity}. In this case, we proposed a novelty loss function called  which combines MAE and MSE. 
\begin{equation}
\begin{split}
    % \mathcal{L}=\sum_{i=1}^D \text{Tanh}(|\Hat{\mathcal{Y}}_i-\mathcal{Y}_i|)|\Hat{\mathcal{Y}}_i-\mathcal{Y}_i|+(1-\text{Tanh}(|\Hat{\mathcal{Y}}_i-\mathcal{Y}_i|))\parallel{\Hat{\mathcal{Y}}_i-\mathcal{Y}_i}\parallel
    %\alpha&=\text{Tanh}(|\Hat{\mathcal{Y}}_i-\mathcal{Y}_i|),\\
    \mathcal{L}&=\sum_{i=1}^D \alpha|\Hat{\mathcal{Y}}_i-\mathcal{Y}_i|+(1-\alpha)\|{\Hat{\mathcal{Y}}_i-\mathcal{Y}_i}{\|}_2, \\
 \alpha&=\text{Tanh}(|\Hat{\mathcal{Y}}_i-\mathcal{Y}_i|),\\
\end{split}
\label{eqn:loss}
\end{equation}
where $\alpha$ is to control the weights of L1 loss and L2 loss. $\alpha$ is based on the absolute error and uses Tanh function \cite{fan2000extended,ghosh2017robust} to map the the absolute error into $(0,1)$ as the control weights. 
The L1 loss plays a prominent role when the prediction error is large.  The L2 loss also contributes when the error is small. 
% In this scenario,  MAE loss assumes a prominent role when the magnitude of the error is substantial, while MSE loss also contributes when the error is minor. The TSM loss has the robustness of MAE and also gives appropriate attention to outliers, thereby enhancing both robustness and accuracy in comparison to MSE loss.

\section{Experiments}
% \subsection{Long-term time series forecasting}
\paragraph{Datasets}
We evaluate TFDNet performance on 8 widely used datasets, including Traffic, Electricity, ILI, and 4 ETT datasets (ETTh1, ETTh2, ETTm1, ETTm2) \cite{wu2021autoformer}. The datasets cover four diverse application domains including energy, traffic, weather, and disease. Following previous studies \cite{wu2021autoformer}, we split these datasets into training, validation, and test sets in chronological order, with the ratio  of 6:2:2 for the ETT dataset and 7:1:2 for the remaining datasets. More details about the datasets are given in Appendix \ref{app:datasets}.

\textbf{Baselines and experimental settings} 
We compare our proposed method with both channel-mixture (CM) models and channel-independence (CI) models. Three state-of-the-art CM models are used as baselines, including FEDformer \cite{zhou2022fedformer}, Autoformer \cite{wu2021autoformer}, and Informer \cite{zhou2022film}. And three CI models including FiLM \cite{zhou2022film}, DLinear \cite{zeng2022transformers}, and PatchTST \cite{nie2022time} are used as baselines. Following the experiment settings of Informer \cite{zhou2021informer} and FiLM \cite{zhou2022film}, we test various lengths of the historical time series with $\inputlen\in\{96, 192, 336, 512, 720\}$. The prediction horizons $\predlen$ are fixed with $T \in \{24, 36, 48, 60\}$ for ILI dataset and $T \in \{96, 192, 336, 720\}$ for the remaing datasets. The commonly used evaluation metrics mean square error (MSE) and mean absolute error (MAE) are used for evaluation. All experiments are repeated 3 times. The details of the implementation setting are in Appendix \ref{app:implementation}.

\textbf{Model variants}
We have two versions of TFDNet, respectively TFDNet-IK and TFDNet-MK. TFDNet-IK is the model variation with the individual kernel strategy (Seasonal-TFB-IK) in Section~\ref{sec:TFB} while TFDNet-MK is the version with the multiple kernel sharing strategy (Seasonal-TFB-MK). The STFT windows lengths $\windowsize$ are set as $\{8, 16, 32\}$ for all datasets and with the strides $\frameshift=\{4, 8, 16\}$. The results from the two versions are reported in Table~\ref{tab:multi} and Table~\ref{tab:uni}.\footnotetext[1]{The initial test data loader implemented by Autoformer discards the final incomplete batch, thereby potentially leading to an overestimation of results. All results reported in this paper are obtained after correcting this issue.}

% \ziyu{move to the implementation details in appendix?TFD-IK employs STFB-IK with individual parameters across channels in Seasonal Encoder, which uses the individual factor $I=D$ for datasets with few channels (ETT, Weather, and ILI) and $I=64$ for datasets with large channels (Traffic and Electricity). TFD-MK utilizes STFB-MK with shared parameters across channels in Seasonal Encoder, which use the number of kernels $G=4$ for datasets with few channels (ETT, Weather, and ILI) and $G=16$ for datasets with large channels (Traffic and Electricity) considering the balance between performance and efﬁciency. The STFT windows lengths $\windowsize$ are set as $\{8, 16, 32\}$ for all datasets and with the strides $\frameshift=\{4, 8, 16\}$}.

%\resizebox{!}{\.5\paperheight}{
% \vskip -0.2in
\begin{table*}[ht]
\centering
%\begin{footnotesize}
% \begin{adjustwidth}{-1.5in}{-1in}
\caption{Multivariate long-term series forecasting results on eight datasets with various prediction lengths $T \in \{96,192,336,720\}$ (For ILI dataset, we set prediction length $T \in \{24,36,48,60\}$). \textbf{Bold}/\underline{underline} indicates the best/second.}
\scalebox{0.6}{
\vspace{-1mm}
% \scalebox{0.60}{
\begin{tabular}{c|c|cccccccccccccccccc}
\toprule
\multicolumn{2}{c|}{Methods}&\multicolumn{2}{c|}{TFDNet-IK}&\multicolumn{2}{c|}{TFDNet-MK}&\multicolumn{2}{c|}{PatchTST}&\multicolumn{2}{c|}{FiLM}&\multicolumn{2}{c|}{DLinear}&\multicolumn{2}{c|}{FEDformer}&\multicolumn{2}{c|}{Autoformer}&\multicolumn{2}{c}{Informer}\\
\midrule
\multicolumn{2}{c|}{Metric} & MSE  & MAE & MSE & MAE& MSE  & MAE& MSE  & MAE& MSE  & MAE& MSE  & MAE& MSE  & MAE & MSE  & MAE\\
\midrule
\multirow{4}{*}{\rotatebox{90}{ETTh1}}       & 96  & \underline{0.359}    & \underline{ 0.386}    & \textbf{0.356} & \textbf{0.383} & 0.373          & 0.402       & 0.377       & 0.401      & 0.375        & 0.396        & 0.376         & 0.416         & 0.449          & 0.455         & 0.916         & 0.741        \\
                             & 192 & \underline{0.398}    & \underline{0.412}    & \textbf{0.396} & \textbf{0.409} & 0.411          & 0.428       & 0.419       & 0.428      & 0.428        & 0.437        & 0.422         & 0.445         & 0.468          & 0.465         & 0.975         & 0.765        \\
                             & 336 & \underline{0.432}    & \underline{0.431}    & \textbf{0.431} & \textbf{0.429} & \underline{0.432}    & 0.445       & 0.466       & 0.466      & 0.448        & 0.449        & 0.452         & 0.463         & 0.506          & 0.494         & 1.099         & 0.819        \\
                             & 720 & \underline{0.438}    & \underline{0.453}    & \textbf{0.421} & \textbf{0.443} & 0.455          & 0.473       & 0.499       & 0.512      & 0.505        & 0.514        & 0.483         & 0.496         & 0.516          & 0.513         & 1.191         & 0.865        \\
\midrule
\multirow{4}{*}{\rotatebox{90}{ETTh2}}       & 96  & \underline{0.268}    & \underline{0.329}    & \textbf{0.266} & \textbf{0.328} & 0.275          & 0.338       & 0.280       & 0.343      & 0.296        & 0.360        & 0.343         & 0.385         & 0.372          & 0.406         & 2.766         & 1.321        \\
                             & 192 & \underline{0.332}    & \underline{0.370}    & \textbf{0.331} & \textbf{0.369} & 0.338          & 0.380       & 0.345       & 0.390      & 0.391        & 0.423        & 0.429         & 0.438         & 0.441          & 0.442         & 6.428         & 2.116        \\
                             & 336 & \textbf{0.359} & \textbf{0.393} & \underline{0.361}    & \underline{0.394}    & 0.366          & 0.403       & 0.372       & 0.415      & 0.445        & 0.460        & 0.489         & 0.485         & 0.479          & 0.478         & 4.779         & 1.833        \\
                             & 720 & \textbf{0.381} & \underline{0.417}    & \textbf{0.381} & \textbf{0.416} & 0.391          & 0.431       & 0.438       & 0.455      & 0.700        & 0.592        & 0.463         & 0.481         & 0.488          & 0.496         & 3.984         & 1.696        \\
\midrule
\multirow{4}{*}{\rotatebox{90}{ETTm1}}       & 96  & \textbf{0.283} & \textbf{0.330} & \underline{0.286}    & \underline{0.333}    & 0.288          & 0.341       & 0.303       & 0.345      & 0.301        & 0.344        & 0.356         & 0.406         & 0.498          & 0.474         & 0.624         & 0.565        \\
                             & 192 & \textbf{0.326} & \textbf{0.355} & \underline{0.327}    & \underline{0.356}    & 0.332          & 0.370       & 0.342       & 0.369      & 0.336        & 0.366        & 0.391         & 0.424         & 0.586          & 0.513         & 0.777         & 0.661        \\
                             & 336 & \textbf{0.359} & \textbf{0.377} & \underline{0.360}    & \underline{0.379}    & 0.362          & 0.392       & 0.371       & 0.387      & 0.371        & 0.387        & 0.441         & 0.453         & 0.657          & 0.543         & 1.087         & 0.799        \\
                             & 720 & \underline{0.412}    & \textbf{0.408} & \textbf{0.410} & \textbf{0.408} & 0.416          & 0.419       & 0.430       & 0.416      & 0.426        & 0.422        & 0.482         & 0.476         & 0.719          & 0.572         & 1.041         & 0.764        \\
\midrule
\multirow{4}{*}{\rotatebox{90}{ETTm2}}        & 96  & \textbf{0.157} & \textbf{0.244} & \underline{ 0.158}    & \underline{0.246}    & 0.162          & 0.254       & 0.167       & 0.257      & 0.170        & 0.264        & 0.189         & 0.281         & 0.254          & 0.325         & 0.427         & 0.503        \\
                             & 192 & \textbf{0.213} & \textbf{0.282} & \underline{0.214}    &\underline{0.283}    & 0.217          & 0.293       & 0.219       & 0.293      & 0.233        & 0.311        & 0.257         & 0.324         & 0.280          & 0.336         & 0.812         & 0.700        \\
                             & 336 & \textbf{0.264} & \textbf{0.318} & \textbf{0.264} &\underline{0.319}    & 0.267          & 0.326       & 0.273       & 0.331      & 0.300        & 0.358        & 0.325         & 0.364         & 0.350          & 0.382         & 1.326         & 0.873        \\
                             & 720 & \textbf{0.345} & \textbf{0.371} & \underline{0.347}    &\underline{0.372}    & 0.353          & 0.382       & 0.356       & 0.387      & 0.422        & 0.439        & 0.429         & 0.424         & 0.437          & 0.429         & 3.986         & 1.532        \\
\midrule
\multirow{4}{*}{\rotatebox{90}{Electricity}}  & 96  & \textbf{0.128} & \textbf{0.221} & \underline{0.129}    & \textbf{0.221} & 0.129          & 0.223       & 0.154       & 0.248      & 0.140        & 0.237        & 0.189         & 0.305         & 0.199          & 0.314         & 0.333         & 0.415        \\
                             & 192 & \textbf{0.145} & \textbf{0.237} & \underline{0.147}    & \underline{0.239}    & 0.149          & 0.243       & 0.167       & 0.260      & 0.154        & 0.250        & 0.205         & 0.320         & 0.215          & 0.325         & 0.344         & 0.427        \\
                             & 336 & \textbf{0.160} & \textbf{0.253} & \underline{0.163}    & \underline{0.256}    & 0.165          & 0.260       & 0.189       & 0.285      & 0.169        & 0.268        & 0.212         & 0.327         & 0.234          & 0.340         & 0.356         & 0.437        \\
                             & 720 & \textbf{0.197} & \textbf{0.285} & \underline{0.200}    & \underline{0.288}    & \underline{0.200}    & 0.292       & 0.250       & 0.341      & 0.204        & 0.300        & 0.245         & 0.352         & 0.289          & 0.380         & 0.386         & 0.450        \\
\midrule
\multirow{4}{*}{\rotatebox{90}{Traffic}}      & 96  & \underline{0.377}    & \underline{0.256}    & \textbf{0.354} & \textbf{0.241} & 0.383          & 0.272       & 0.413       & 0.290      & 0.412        & 0.286        & 0.577         & 0.360         & 0.642          & 0.408         & 0.735         & 0.415        \\
                             & 192 & 0.392          & 0.263          & \textbf{0.372} & \textbf{0.250} & \underline{0.380}    & \underline{0.259} & 0.409       & 0.289      & 0.424        & 0.291        & 0.607         & 0.374         & 0.640          & 0.405         & 0.749         & 0.422        \\
                             & 336 & \underline{0.406}    & \underline{0.266}    & \textbf{0.388} & \textbf{0.257} & 0.410          & 0.285       & 0.425       & 0.299      & 0.438        & 0.299        & 0.624         & 0.384         & 0.621          & 0.384         & 0.855         & 0.486        \\
                             & 720 & \underline{0.447}    & \underline{0.287}    & \textbf{0.428} & \textbf{0.279} & 0.454          & 0.313       & 0.525       & 0.373      & 0.467        & 0.317        & 0.625         & 0.381         & 0.709          & 0.435         & 1.094         & 0.619        \\
\midrule
\multirow{4}{*}{\rotatebox{90}{Weather}}    & 96  & \textbf{0.143} & \textbf{0.188} & \underline{0.148}    & \underline{0.194}    & \underline{0.148}    & 0.200       & 0.194       & 0.234      & 0.175        & 0.235        & 0.221         & 0.304         & 0.271          & 0.341         & 0.382         & 0.437        \\
                             & 192 & \textbf{0.186} & \textbf{0.230} & 0.192          & \underline{0.236}    & \underline{0.191}    & 0.241       & 0.229       & 0.266      & 0.216        & 0.274        & 0.325         & 0.372         & 0.320          & 0.374         & 0.566         & 0.522        \\
                             & 336 & \textbf{0.236} & \textbf{0.269} & 0.243          & \underline{0.275}    & \underline{0.240}    & 0.281       & 0.266       & 0.295      & 0.262        & 0.314        & 0.386         & 0.408         & 0.350          & 0.387         & 0.614         & 0.555        \\
                             & 720 & \underline{0.314}    & \textbf{0.326} & 0.319          & 0.331          & \textbf{0.307} & \underline{0.329} & 0.323       & 0.340      & 0.327        & 0.367        & 0.415         & 0.423         & 0.428          & 0.434         & 1.098         & 0.760        \\
\midrule
\multirow{4}{*}{\rotatebox{90}{ILI}}         & 24  & \underline{1.834}    & \textbf{0.823} & \textbf{1.786} & \underline{0.829}    & 2.123          & 0.920       & 2.297       & 0.957      & 2.260        & 1.052        & 3.232         & 1.266         & 3.523          & 1.313         & 5.219         & 1.556        \\
                             & 36  & \underline{1.780}    & \textbf{0.834} & \textbf{1.764} & \underline{0.836}    & 1.877          & 0.934       & 2.298       & 0.976      & 2.258        & 1.061        & 3.122         & 1.218         & 3.359          & 1.264         & 5.082         & 1.562        \\
                             & 48  & 1.815          & \underline{0.861}    & \textbf{1.701} & \textbf{0.844} & \underline{1.739}    & 0.885       & 2.344       & 1.025      & 2.321        & 1.087        & 2.953         & 1.175         & 3.358          & 1.252         & 5.177         & 1.573        \\
                             & 60  & \textbf{1.756} & \textbf{0.861} & 1.910          & 0.936          & \underline{1.808}    & \underline{0.914} & 2.151       & 0.952      & 2.478        & 1.130        & 2.858         & 1.159         & 2.892          & 1.150         & 5.280         & 1.584  \\
\bottomrule
\end{tabular}
\label{tab:multi}
}
\end{table*}
%\resizebox{!}{\.5\paperheight}{
% \vskip -0.2in
\begin{table*}[ht]
\centering
%\begin{footnotesize}
% \begin{adjustwidth}{-1.5in}{-1in}
\caption{Univariate long-term series forecasting results on ETT datasets with various prediction lengths $T \in \{96,192,336,720\}$. \textbf{Bold}/\underline{underline} indicates the best/second.}
\label{tab:uni}
\scalebox{0.60}{
\vspace{-1mm}
% \scalebox{0.60}{
\begin{tabular}{c|c|cccccccccccccccccc}
\toprule
\multicolumn{2}{c|}{Methods}                      & \multicolumn{2}{c|}{TFDNet-MK}  & \multicolumn{2}{c|}{PatchTST}   & \multicolumn{2}{c|}{FiLM} & \multicolumn{2}{c|}{Dlinear} & \multicolumn{2}{c|}{FEDformer} & \multicolumn{2}{c|}{Autoformer} & \multicolumn{2}{c}{Informer} \\ \midrule
\multicolumn{2}{c|}{Metric}                       & MSE            & MAE            & MSE            & MAE            & MSE         & MAE         & MSE          & MAE          & MSE           & MAE           & MSE            & MAE           & MSE           & MAE          \\ \midrule
\multicolumn{1}{c|}{\multirow{4}{*}{{\rotatebox{90}{ETTh1}}}} & 96  & \textbf{0.054} & \textbf{0.176} & \textbf{0.054} & \underline{0.177}    & 0.057       & 0.182       & 0.056        & 0.180        & 0.084         & 0.220         & 0.088          & 0.238         & 0.172         & 0.350        \\
\multicolumn{1}{c|}{}                       & 192 & \textbf{0.069} & \textbf{0.202} & \underline{0.071}    & \underline{0.205}    & 0.072       & 0.207       & 0.075        & 0.209        & 0.108         & 0.249         & 0.097          & 0.239         & 0.280         & 0.461        \\
\multicolumn{1}{c|}{}                       & 336 & \textbf{0.077} & \textbf{0.221} & \underline{0.083}    & \underline{0.228}    & 0.084       & 0.230       & 0.092        & 0.238        & 0.123         & 0.278         & 0.104          & 0.260         & 0.256         & 0.428        \\
\multicolumn{1}{c|}{}                       & 720 & \underline{0.090}    & \underline{0.237}    & \textbf{0.086} & \textbf{0.234} & 0.091       & 0.239       & 0.168        & 0.334        & 0.146         & 0.303         & 0.132          & 0.290         & 0.185         & 0.356        \\ \midrule
\multicolumn{1}{c|}{\multirow{4}{*}{{\rotatebox{90}{ETTh2}}}} & 96  & \textbf{0.126} & \textbf{0.274} & \underline{0.130}    & \underline{0.283}    & 0.128       & 0.273       & 0.132        & 0.280        & 0.129         & 0.271         & 0.152          & 0.303         & 0.221         & 0.381        \\
\multicolumn{1}{c|}{}                       & 192 & \textbf{0.163} & \textbf{0.319} & \underline{0.169}    & \underline{0.328}    & 0.189       & 0.343       & 0.176        & 0.329        & 0.187         & 0.330         & 0.191          & 0.340         & 0.280         & 0.431        \\
\multicolumn{1}{c|}{}                       & 336 & \textbf{0.188} & \textbf{0.351} & \underline{0.193}    & \underline{0.358}    & 0.201       & 0.364       & 0.210        & 0.369        & 0.231         & 0.378         & 0.233          & 0.380         & 0.323         & 0.456        \\
\multicolumn{1}{c|}{}                       & 720 & \textbf{0.216} & \textbf{0.373} & \underline{0.221}    & \underline{0.379}    & 0.224       & 0.380       & 0.290        & 0.438        & 0.278         & 0.422         & 0.260          & 0.404         & 0.293         & 0.440        \\ \midrule
\multicolumn{1}{c|}{\multirow{4}{*}{{\rotatebox{90}{ETTm1}}}} & 96  & \textbf{0.026} & \textbf{0.121} & \textbf{0.026} & \textbf{0.121} & 0.029       & 0.128       & 0.028        & 0.125        & 0.034         & 0.143         & 0.050          & 0.174         & 0.103         & 0.264        \\
\multicolumn{1}{c|}{}                       & 192 & \textbf{0.039} & \textbf{0.150} & \textbf{0.039} & \textbf{0.150} & 0.041       & 0.154       & 0.043        & 0.154        & 0.066         & 0.203         & 0.110          & 0.250         & 0.221         & 0.404        \\
\multicolumn{1}{c|}{}                       & 336 & \textbf{0.052} & \textbf{0.173} & \underline{0.053}    & \textbf{0.173} & 0.053       & 0.174       & 0.062        & 0.183        & 0.071         & 0.210         & 0.085          & 0.236         & 0.219         & 0.387        \\
\multicolumn{1}{c|}{}                       & 720 & \textbf{0.070} & \textbf{0.200} & 0.073          & 0.206          & \underline{0.071} & \underline{0.205} & 0.080        & 0.211        & 0.109         & 0.259         & 0.120          & 0.283         & 0.456         & 0.601        \\ \midrule
\multicolumn{1}{c|}{\multirow{4}{*}{{\rotatebox{90}{ETTm2}}}} & 96  & \textbf{0.062} & \textbf{0.182} & \underline{0.065}    & \underline{0.186}    & 0.066       & 0.191       & 0.064        & 0.184        & 0.066         & 0.197         & 0.098          & 0.241         & 0.076         & 0.210        \\
\multicolumn{1}{c|}{}                       & 192 & \textbf{0.091} & \textbf{0.224} & 0.094          & 0.231          & 0.096       & 0.235       & \underline{0.092}  & \underline{0.227}  & 0.104         & 0.250         & 0.164          & 0.315         & 0.124         & 0.274        \\
\multicolumn{1}{c|}{}                       & 336 & \textbf{0.117} & \textbf{0.260} & \underline{0.120}    & \underline{0.265}    & 0.123       & 0.269       & 0.122        & 0.265        & 0.155         & 0.303         & 0.178          & 0.325         & 0.163         & 0.314        \\
\multicolumn{1}{c|}{}                       & 720 & \textbf{0.170} & \textbf{0.321} & \underline{0.172}    & \underline{0.322}    & 0.173       & 0.323       & 0.174        & 0.320        & 0.193         & 0.343         & 0.189          & 0.340         & 0.293         & 0.431        \\ \bottomrule

\end{tabular}
}
\end{table*}

\subsection{Main results}
% Table \ref{tab:multi-benchmarks} shows the multivariate long-term forecasting results. 
\paragraph{Multivariate results} 
From Table~\ref{tab:multi}, we can see that our models outperform all baselines. In comparison with the best CM model FEDformer, TFDNet-IK with the CI strategy achieves an overall reduction of $32.3\%$ in MSE and $22.4\%$ in MAE, and TFDNet-MK achieves a reduction of $32.7\%$ and $22.0\%$, respectively. 
%Due to the time-frequency enhanced structure and mixture loss design, TFDNets also exceed PatchTST, the best CI model available, on all datasets. 
Compared with the best CI model PatchTST, TFDNet-IK can achieve a reduction of $3.0\%$ in MSE and $4.7\%$ in MAE, while TFDNet-MK obtains a reduction of $3.6\%$ and $4.2\%$. 
The two versions of TFDNet demonstrate different performances on varying datasets and validate that the channel-wise correlations have important effects on the design of kernel operations (individual vs. sharing). The TFDNet-MK model has superior performance on datasets with higher channel correlations in the seasonal component, such as Traffic. Conversely, TFDNet-IK has better performance on datasets with lower channel correlations, such as ETTm1, ETTm2, and Weather. 

%Like other CI models, TFDNets outperform all CM models significantly. TFDNets also outperform the best CI model PatchTST on all datasets with different settings except for horizon 720 in Weather. 
\textbf{Univariate results} We present the univariate forecasting results on the four ETT datasets in Table \ref{tab:uni}. Similarly, TFDNet-MK has superior performance compared to all other baseline methods. Due to its single channel, TDF-IK with individual kernels is not applicable.
% \usepackage{booktabs}
% \usepackage{multirow}
\begin{table*}[h]
\centering
\caption{Ablation of decomposition structure and time-frequency enhanced encoder. The TFNet-T utilizes only TrendEncoder without decomposition. FreqNet and TimeNet process time series in the single frequency domain and single time domain, respectively. \textbf{Bold} indicates the best.}
\label{ablation}
\vspace{-1mm}
\scalebox{0.6}{
\begin{tabular}{@{}cc|llll|llll|llll|llll@{}}
\toprule
\multicolumn{1}{l}{Datasets}                    & \multicolumn{1}{l|}{} & \multicolumn{4}{c|}{Traffic}                                                                          & \multicolumn{4}{c|}{Electricity}                                                                      & \multicolumn{4}{c|}{Weather}                                                                          & \multicolumn{4}{c}{ETTm2}                                                                            \\ \midrule
\multicolumn{2}{l|}{Prediction Length T}                                & \multicolumn{1}{c}{96} & \multicolumn{1}{c}{192} & \multicolumn{1}{c}{336} & \multicolumn{1}{c|}{720} & \multicolumn{1}{c}{96} & \multicolumn{1}{c}{192} & \multicolumn{1}{c}{336} & \multicolumn{1}{c|}{720} & \multicolumn{1}{c}{96} & \multicolumn{1}{c}{192} & \multicolumn{1}{c}{336} & \multicolumn{1}{c|}{720} & \multicolumn{1}{c}{96} & \multicolumn{1}{c}{192} & \multicolumn{1}{c}{336} & \multicolumn{1}{c}{720} \\ \midrule
\multicolumn{1}{c|}{\multirow{2}{*}{TFDNet-IK}} & MSE                   & 0.377                  & 0.392                   & 0.406                   & 0.447                    & \textbf{0.128}         & \textbf{0.145}          & \textbf{0.160}          & \textbf{0.197}           & \textbf{0.143}         & \textbf{0.186}          & \textbf{0.236}          & \textbf{0.314}           & \textbf{0.157}         & \textbf{0.213}          & \textbf{0.264}          & \textbf{0.345}          \\
\multicolumn{1}{c|}{}                           & MAE                   & 0.256                  & 0.263                   & 0.266                   & 0.287                    & \textbf{0.221}         & \textbf{0.237}          & \textbf{0.253}          & \textbf{0.285}           & \textbf{0.188}         & \textbf{0.230}          & \textbf{0.269}          & \textbf{0.326}           & \textbf{0.244}         & \textbf{0.282}          & \textbf{0.318}          & \textbf{0.371}          \\ \midrule
\multicolumn{1}{c|}{\multirow{2}{*}{TFDNet-MK}} & MSE                   & \textbf{0.354}         & \textbf{0.372}          & \textbf{0.388}          & \textbf{0.428}           & 0.129                  & 0.147                   & 0.163                   & 0.200                    & 0.148                  & 0.192                   & 0.243                   & 0.319                    & 0.158                  & 0.214                   & \textbf{0.264 }                  & 0.347                   \\
\multicolumn{1}{c|}{}                           & MAE                   & \textbf{0.241}         & \textbf{0.250}          & \textbf{0.257}          & \textbf{0.279}           & \textbf{0.221}         & 0.239                   & 0.256                   & 0.288                    & 0.194                  & 0.236                   & 0.275                   & 0.331                    & 0.246                  & 0.283                   & 0.319                   & 0.372                   \\ \midrule
\multicolumn{1}{c|}{\multirow{2}{*}{TFNet-T}}  & MSE                   & 0.372                  & 0.391                   & 0.405                   & 0.442                    & 0.130                  & 0.146                   & 0.163                   & 0.200                    & 0.147                  & 0.192                   & 0.243                   & 0.318                    & 0.159                  & 0.214                   & 0.266                   & 0.348                   \\
\multicolumn{1}{c|}{}                           & MAE                   & 0.252                  & 0.260                   & 0.266                   & 0.287                    & 0.223                  & 0.238                   & 0.255                   & 0.288                    & 0.193                  & 0.235                   & 0.275                   & 0.329                    & 0.246                  & 0.284                   & 0.319                   & 0.373                   \\ \midrule
\multicolumn{1}{c|}{\multirow{2}{*}{FreqNet}}  & MSE                   & 0.385                  & 0.396                   & 0.409                   & 0.446                & 0.136                  & 0.149                   & 0.165                   & 0.204                    & 0.165                  & 0.207                   & 0.256                   & 0.322                    & 0.158                  & 0.214                   & 0.265                   & 0.351                   \\
\multicolumn{1}{c|}{}                           & MAE                   & 0.260                  & 0.264                   & 0.270                   &0.290                          & 0.229         & 0.240                   & 0.257                   & 0.290                    & 0.212                  & 0.249                   & 0.286                   & 0.334                    & 0.246                  & 0.287                   & 0.320                   & 0.376                   \\ \midrule
\multicolumn{1}{c|}{\multirow{2}{*}{TimeNet}} & MSE                   & 0.393                  & 0.403                   & 0.415                   & 0.453                    & 0.136                  & 0.148                   & 0.164                   & 0.203                    & 0.169                  & 0.212                   & 0.258                   & 0.326                    & 0.159                  & 0.215                   & 0.268                   & 0.353                   \\
\multicolumn{1}{c|}{}                           & MAE                   & 0.262                  & 0.265                   & 0.271                   & 0.291                    & 0.230                  & 0.240                   & 0.257                   & 0.289                    & 0.215                  & 0.252                   & 0.288                   & 0.338                    & 0.247                  & 0.286                   & 0.322                   & 0.376                   \\ \bottomrule
\end{tabular}
}
\end{table*}
\subsection{Ablation studies}
%\ziyu{merge Table 4 and Table 5}
% \input{Table/ab_decompose}

To inspect the effects of some main components, we perform ablation studies by removing the decomposition structure and investigating different variants of time-frequency enhanced encoders (FreqNet and TimeNet). The ablation results are shown in Table~\ref{ablation}.

\paragraph{Ablation of decomposition structure}
TFNet-T only keeps TrendEncoder with Trend-TFB to process the original data, removing the decomposition process. We can see that the performance of TFNet-T drops, which indicates the effects of Seaonal-TFB with individual kernels or multiple kernels in the decomposition structure. %\ziyu{some did not drop?}

% We also compared TFNet-T with a linear regression model to prove the validity of our proposed encoder structure with TFB and FFFN. As shown in Table?  
% \input{Table/ab_tf_structure}

\paragraph{Ablation of time-frequency enhanced encoder}
We investigate the effects of different operations in the time-frequency enhanced encoder. We have two variants FreqNet and TimeNet. 
FreqNet only utilizes the information from the frequency domain through Frequency-FFN. TimeNet replaces the encoder with two linear layers to learn the representation from the time domain in the seasonal and trend branches.
%\ziyu{how about only removing TFB?, how about removing multi-scale?}
We can see that the performance of the two variants remarkably drops. It verifies the effects of the time-frequency enhanced operations in the time-frequency enhanced encoder.

% \usepackage{multirow}
\begin{table*}[ht]
\centering
\caption{Memory usages and running time per epoch of PatchTST, FiLM, and TFDNet. The "-" indicates the out-of-memory. \textbf{Bold} indicates the best.}
\scalebox{0.8}{
\begin{tabular}{c|ccccc|ccccc}
\toprule
\multicolumn{1}{c|}{Methods}&\multicolumn{5}{c|}{Memory Usage (MB)} & \multicolumn{5}{c}{Runing Time Per Epoch (s)} \\ \midrule
\multicolumn{1}{c|}{Input Length $L$}                               & 192      & 336     & 720     & 1440    & 2880    & 192      & 336       & 720       & 1440      & 2880 \\ \midrule
PatchTST                                               & 3708     & 5292    & 13892   & 43866   & -       & 200      & 305       & 635       & 1879      & -        \\ 
FiLM                                                   & 10300    & 16834   & 34338   & -       & -       & 715      & 1159      & 2302      & -         & -        \\ 
TFDNet-IK                                              & \textbf{712}      & \textbf{866}     & \textbf{2046}    & 6272    & 22722   & \textbf{177}      & \textbf{190}       & \textbf{222}       & \textbf{405}       & 949      \\ 
TFDNet-MK                                              & 1106     & 1196    & 2524    & \textbf{4676}    & \textbf{10192}   & 188      & 191       & 249       & 416       & \textbf{818}      \\ 
\bottomrule
\end{tabular}
}
\label{effenciency}
\end{table*}

\subsection{Model effenciency}
Running efficiency plays a crucial role in time series forecasting. We utilize the Electricity dataset as an example to investigate the model efficiency (speed and memory usage). In the Electricity dataset, the channel number is 321 and the batch size is set as 8. The prediction horizon $\predlen$  is fixed at 720 and we increase the look-back window $\inputlen$ from $192$ to $2880$. %containing 321 channels of batch size 8 to evaluate the training speed and memory usage on the modeling process, compared with the other two channel-independence models FiLM and PatchTST.We fix the prediction horizons $\predlen$ at 720 and increase the look-back window $\inputlen$ from $192$ to $2880$. 
All results are running on a single NVIDIA RTX A6000 48GB GPU.
%with the same dataloader provided by \cite{wu2021autoformer}. % \cite{nie2022time} \cite{zhou2022film}
Compared with other channel-independence models FiLM and PatchTST, our proposed TFDNets achieve higher efficiency by reducing the model complexity in terms of both memory usage and training speed per epoch. The computational complexity of the TFB kernel operation is $\mathcal{O}(MN^2)$ ($N \ll 
 L, M \ll L$). When the input length scales up, we reduce memory usage and running time with up to $9.4\times$ and $4.63\times$, respectively. When $\inputlen$ is very large, e.g. $1440$ or $2880$, the compared models cannot run due to out of max memory usage of a single GPU while our method is scaleable with the large input length.

 % \subsection{Prediction Visualization}

% \begin{figure}[!tb]
%   \centering
%   \begin{subfigure}[b]{0.46\textwidth}
%     \includegraphics[width=\textwidth]{graphs/memory_useage.png}
%   \end{subfigure}
%   \hspace{0.02\textwidth} 
%   \begin{subfigure}[b]{0.46\textwidth}
%     \includegraphics[width=\textwidth]{graphs/running_time.png}
%   \end{subfigure}
%   \caption{(Left) the memory usage of TFDNet and baseline models. (Right) training speed per epoch of TFNet and baseline models.}
%   \label{model_effciency}
% \end{figure}

\section{Conclusion and limitation}
% The paper proposes an effective design of long-term time series forecasting methods through the time-frequency domain by introducing time-frequency blocks. Besides, TFDNet utilizes the seasonal-trend decomposition to process the two components considering not only the difference between the two parts but also the variation of channel correlations. Our model exhibits the validity of representation in the time-frequency domain and the channel individual setting through extensive experiments. TFDNet can yield consistent state-of-the-art performance on real-world datasets with high efficiency.
We take a holistic view to process long-term time series from the time-frequency domain and propose TFDNet by leveraging Short-time Fourier transform (STFT) to transform long-term time series into time-frequency matrixes. And multi-scale time-frequency enhanced encoders are devised, in which two separate time-frequency blocks (TFBs) are respectively developed for the trend component and the seasonal component to extract the distinct underlying patterns within the two components. The observation of different channel-wise correlation patterns is integrated into kernel operations of time-frequency blocks and two kernel learning strategies (individual vs. sharing) are explored to deal with various channel-wise correlation patterns in the seasonal component. Experiment results validate the effectiveness and efficiency of TFDNet. Although the two versions of TFDNet have achieved superior performances, the model selection is mainly based on the channel-wise correlation effects. Further investigations and theoretical studies should be considered. 
%In Future work, different levels of channel-wise correlation effects can be further explored.

\newpage
\bibliography{paper}
\bibliographystyle{neurips_2022}

% \section*{References}

% References follow the acknowledgments in the camera-ready paper. Use unnumbered first-level heading for
% the references. Any choice of citation style is acceptable as long as you are
% consistent. It is permissible to reduce the font size to \verb+small+ (9 point)
% when listing the references.
% Note that the Reference section does not count towards the page limit.
% \medskip

% {
% \small

% [1] Alexander, J.A.\ \& Mozer, M.C.\ (1995) Template-based algorithms for
% connectionist rule extraction. In G.\ Tesauro, D.S.\ Touretzky and T.K.\ Leen
% (eds.), {\it Advances in Neural Information Processing Systems 7},
% pp.\ 609--616. Cambridge, MA: MIT Press.

% [2] Bower, J.M.\ \& Beeman, D.\ (1995) {\it The Book of GENESIS: Exploring
%   Realistic Neural Models with the GEneral NEural SImulation System.}  New York:
% TELOS/Springer--Verlag.

% [3] Hasselmo, M.E., Schnell, E.\ \& Barkai, E.\ (1995) Dynamics of learning and
% recall at excitatory recurrent synapses and cholinergic modulation in rat
% hippocampal region CA3. {\it Journal of Neuroscience} {\bf 15}(7):5249-5262.
% }

% %%%%%%%%%%%%%%%%%%%%%%%%%%%%%%%%%%%%%%%%%%%%%%%%%%%%%%%%%%%%

\newpage
\appendix
\section{Datasets and Implementation Details}
\label{app:section_dataset_implem}
\subsection{Dataset Details}
\label{app:datasets}
This subsection provides a summary of the datasets utilized in this paper:
1) ETT\footnote{https://github.com/zhouhaoyi/ETDataset} (Electricity Transformer Temperature) dataset contains two electric transformers, ETT1 and ETT2, collected from two separate counties. Each of them has two versions of sampling resolutions (15min \& 1h). Thus, there are four ETT datasets: ETTm1, ETTm2, ETTh1, and ETTh2.
2) Electricity\footnote{https://archive.ics.uci.edu/ml/datasets/ElectricityLoadDiagrams20112014} dataset contains the hourly electricity consumption for 321 customers.
3) Traffic\footnote{http://pems.dot.ca.gov} dataset contains the freeway occupation rate from 862 different sensors in California, USA. 
4) Weather\footnote{https://www.bgc-jena.mpg.de/wetter/} dataset contains 21 meteorological indicators in Germany, such as humidity and air temperature.
5) Illness\footnote{https://gis.cdc.gov/grasp/fluview/fluportaldashboard.html} dataset contains the number of influenza-like illness patients weekly in the United States. 
Table \ref{tab:dataset} summarizes the features of the five benchmark datasets. \begin{table*}[h]
\caption{Details of benchmark datasets.}
\label{tab:dataset}
\begin{center}
\begin{small}
\scalebox{0.85}{
\begin{tabular}{l|cccccccc}
\toprule
% Dataset & num & dim & freq \\
Dataset &ETTm1 & ETTm2 & ETTh1 & ETTh2 & Electricity & Weather & Traffic & ILI\\
\midrule
Length & 69680 & 69680 & 17420 & 17420 & 26304 & 52696 & 17544 & 966\\
Feature & 7 & 7 & 7 & 7 & 321 & 21 & 862 & 7\\
Frequency & 15 min & 15 min & 1 h & 1 h & 1 h & 10 min & 1 h & 7 days\\
\bottomrule
\end{tabular}
}
\end{small}
\end{center}

\end{table*}

The Exchange-rate\footnote{https://github.com/laiguokun/multivariate-time-series-data} dataset is used as the benchmark dataset, similar to PatchTST because the financial datasets exhibit distinct characteristics in comparison to time series datasets in other domains, such as their level of predictability \cite{nie2022time}.

\subsection{Implementation Details}
TFDNet-IK employs Seasonal-TFB-IK with individual parameters across channels in SeasonalEncoder, which uses the individual factor $I=D$ for datasets with few channels (ETT, Weather, and ILI) and $I=64$ for datasets with large channels (Traffic and Electricity). TFDNet-MK utilizes Seasonal-TFB-MK with shared parameters across channels in Seasonal Encoder, which use the number of kernels $k=4$ for datasets with few channels (ETT, Weather, and ILI) and $k=16$ for datasets with large channels (Traffic and Electricity) considering the balance between performance and efﬁciency.

Our model is trained with the proposed Mixture loss in section \ref{loss}, using the Adam \cite{kingma2014adam} optimizer with an initial learning rate from $5e^{-5}$ to $5e^{-3}$. We also tune the learning rate with the cosine decay schedule. The batch size is set to 128 for the small dataset, eg. ETT, and 32 for the large dataset, eg. Traffic due to the limation of GPU memory usage. The default training process is 50 epochs (100 epochs for Electricity and Traffic) within 10 epochs early stopping. We save the best model with the lowest loss on the validation as the ﬁnal testing. All experiments are repeated 3 times, implemented in PyTorch \cite{paszke2019pytorch} and conducted on a single NVIDIA RTX A6000 48GB GPUs which is sufficient for all our experiments.
\label{app:implementation}
\section{Experiment Error Bars}
We train all models 3 times and calculate the error bars for TFDNet and another two SOTA models (PatchTST and FiLM) for evaluating the robustness of models. The results are shown in Table \ref{tab:error_bar}. We can see our proposed TFDNet-IK and TFDNet-MK are able to outperform the other models with lower standard deviation. 
\begin{table*}[tb]
\centering
%\begin{footnotesize}
% \begin{adjustwidth}{-1.5in}{-1in}
\caption{Multivariate long-term series forecasting results with error bars (Mean and STD). All experiments are repeated 3 times. \textbf{Bold} indicates the best.}
\scalebox{0.6}{
\vspace{-1mm}
\begin{tabular}{@{}c|c|cccccccc@{}}
\toprule
\multicolumn{2}{c|}{Methods}                            & \multicolumn{2}{c|}{TFDNet-IK}                  & \multicolumn{2}{c|}{TFDNet-MK}                  & \multicolumn{2}{c|}{PatchTST}          & \multicolumn{2}{c}{FiLM}      \\ \midrule
\multicolumn{2}{c|}{Metric}                             & MSE                    & \multicolumn{1}{c|}{MAE}                    & MSE                    & \multicolumn{1}{c|}{MAE}                    & MSE                    & \multicolumn{1}{c|}{MAE}           & MSE           & MAE           \\ \midrule
\multirow{4}{*}{\rotatebox{90}{ETTh1}}      & 96  & 0.3590±0.0014          & 0.3861±0.0003          & \textbf{0.3560±0.0003} & \textbf{0.3833±0.0002} & 0.3732±0.0009          & 0.4023±0.0008 & 0.3772±0.0021 & 0.4012±0.0024 \\
                            & 192 & 0.3976±0.0015          & 0.4120±0.0010          & \textbf{0.3957±0.0005} & \textbf{0.4090±0.0008} & 0.4111±0.0014          & 0.4277±0.0015 & 0.4193±0.0020 & 0.4283±0.0022 \\
                            & 336 & 0.4319±0.0011          & 0.4310±0.0007          & \textbf{0.4308±0.0011} & \textbf{0.4290±0.0013} & 0.4323±0.0059          & 0.4455±0.0048 & 0.4659±0.0040 & 0.4663±0.0037 \\
                            & 720 & 0.4380±0.0016          & 0.4530±0.0016          & \textbf{0.4208±0.0034} & \textbf{0.4433±0.0033} & 0.4552±0.0049          & 0.4728±0.0038 & 0.4993±0.0055 & 0.5120±0.0039 \\ \midrule
\multirow{4}{*}{\rotatebox{90}{ETTh2}}       & 96  & 0.2675±0.0012          & 0.3291±0.0008          & \textbf{0.2658±0.0002} & \textbf{0.3281±0.0003} & 0.2745±0.0011          & 0.3379±0.0006 & 0.28±0.0005   & 0.3429±0.0003 \\
                            & 192 & 0.3316±0.0010          & 0.3695±0.0006          & \textbf{0.3306±0.0016} & \textbf{0.3687±0.0009} & 0.3383±0.0020          & 0.3798±0.0015 & 0.3452±0.0020 & 0.3895±0.0011 \\
                            & 336 & \textbf{0.3590±0.0012} & \textbf{0.3935±0.0006} & 0.3610±0.0013          & 0.3944±0.0008          & 0.3655±0.0011          & 0.4032±0.0005 & 0.3716±0.0022 & 0.4147±0.0012 \\
                            & 720 & 0.3814±0.0014          & 0.4172±0.0012          & \textbf{0.3809±0.0017} & \textbf{0.4163±0.0019} & 0.3914±0.0021          & 0.4310±0.0016 & 0.4380±0.0039 & 0.4548±0.0038 \\ \midrule
\multirow{4}{*}{\rotatebox{90}{ETTm1}}       & 96  & \textbf{0.2830±0.0010} & \textbf{0.3298±0.0006} & 0.2859±0.0005          & 0.3325±0.0004          & 0.2882±0.0006          & 0.3413±0.0010 & 0.3028±0.0004 & 0.3448±0.0004 \\
                            & 192 & \textbf{0.3264±0.0007} & \textbf{0.3554±0.0005} & 0.3274±0.0001          & 0.3564±0.0004          & 0.3320±0.0009          & 0.3699±0.0007 & 0.3418±0.0014 & 0.3695±0.0006 \\
                            & 336 & \textbf{0.3590±0.0009} & \textbf{0.3767±0.0005} & 0.3595±0.0022          & 0.3786±0.0018          & 0.3617±0.0018          & 0.3922±0.0002 & 0.3712±0.0008 & 0.3871±0.0004 \\
                            & 720 & 0.4121±0.0013          & \textbf{0.4079±0.0005} & \textbf{0.4101±0.0001} & 0.4082±0.0003          & 0.4160±0.0002          & 0.4186±0.0003 & 0.4297±0.0009 & 0.4162±0.0007 \\\midrule
\multirow{4}{*}{\rotatebox{90}{ETTm2}}       & 96  & \textbf{0.1569±0.0006} & \textbf{0.2439±0.0007} & 0.1584±0.0002          & 0.2462±0.0009          & 0.1624±0.0008          & 0.2543±0.0004 & 0.1674±0.0009 & 0.2573±0.0004 \\
                            & 192 & \textbf{0.2129±0.0000} & \textbf{0.2821±0.0001} & 0.2136±0.0002          & 0.2834±0.0002          & 0.2168±0.0009          & 0.2932±0.0006 & 0.2193±0.0002 & 0.2934±0.0003 \\
                            & 336 & \textbf{0.2636±0.0004} & \textbf{0.3176±0.0003} & 0.2638±0.0005          & 0.3186±0.0002          & 0.2670±0.0005          & 0.3263±0.0005 & 0.2727±0.0006 & 0.3305±0.0007 \\
                            & 720 & \textbf{0.3445±0.0011} & \textbf{0.3706±0.0003} & 0.3468±0.0007          & 0.3723±0.0005          & 0.3526±0.0029          & 0.3819±0.0027 & 0.3563±0.0007 & 0.3875±0.0009 \\ \midrule
\multirow{4}{*}{\rotatebox{90}{Electricity}} & 96  & \textbf{0.1277±0.0001} & \textbf{0.2210±0.0001} & 0.1289±0.0003          & 0.2215±0.0004          & 0.1290±0.0005          & 0.2226±0.0005 & 0.1544±0.0000 & 0.2482±0.0002 \\
                            & 192 & \textbf{0.1453±0.0002} & \textbf{0.2372±0.0001} & 0.1466±0.0006          & 0.2386±0.0007          & 0.1489±0.0006          & 0.2427±0.0014 & 0.1669±0.0002 & 0.2599±0.0003 \\
                            & 336 & \textbf{0.1604±0.0004} & \textbf{0.2533±0.0004} & 0.1630±0.0004          & 0.2556±0.0003          & 0.1646±0.0010          & 0.2599±0.0012 & 0.1895±0.0001 & 0.2847±0.0003 \\
                            & 720 & \textbf{0.1965±0.0010} & \textbf{0.2853±0.0006} & 0.2001±0.0005          & 0.2878±0.0004          & 0.2003±0.0027          & 0.2919±0.0020 & 0.2501±0.0003 & 0.3411±0.0002 \\ \midrule
\multirow{4}{*}{\rotatebox{90}{Traffic}}     & 96  & 0.3767±0.0015          & 0.2557±0.0004          & \textbf{0.3542±0.0007} & \textbf{0.2412±0.0008} & 0.3834±0.0182          & 0.2715±0.0174 & 0.4133±0.0012 & 0.2897±0.0001 \\
                            & 192 & 0.3916±0.0001          & 0.2625±0.0001          & \textbf{0.3720±0.0005} & \textbf{0.2496±0.0003} & 0.3798±0.0011          & 0.2587±0.0007 & 0.4087±0.0001 & 0.2888±0.0002 \\
                            & 336 & 0.4063±0.0008          & 0.2657±0.0007          & \textbf{0.3880±0.0008} & \textbf{0.2567±0.0008} & 0.4099±0.0129          & 0.2847±0.0147 & 0.4255±0.0006 & 0.2993±0.0008 \\
                            & 720 & 0.4474±0.0020          & 0.2872±0.0008          & \textbf{0.4285±0.0016} & \textbf{0.2786±0.0019} & 0.4538±0.0018          & 0.3129±0.0014 & 0.5253±0.0009 & 0.3729±0.0002 \\ \midrule
\multirow{4}{*}{\rotatebox{90}{Weather}}     & 96  & \textbf{0.1427±0.0002} & \textbf{0.1883±0.0001} & 0.1479±0.0004          & 0.1941±0.0000          & 0.1481±0.0029          & 0.1997±0.0035 & 0.1938±0.0009 & 0.2344±0.0005 \\
                            & 192 & \textbf{0.1859±0.0003} & \textbf{0.2296±0.0007} & 0.1923±0.0002          & 0.2358±0.0005          & 0.1913±0.0025          & 0.2410±0.0017 & 0.2293±0.0003 & 0.2656±0.0002 \\
                            & 336 & \textbf{0.2356±0.0005} & \textbf{0.2693±0.0004} & 0.2426±0.0004          & 0.2754±0.0002          & 0.2403±0.0010          & 0.2809±0.0018 & 0.2661±0.0003 & 0.2950±0.0004 \\
                            & 720 & 0.3138±0.0019          & \textbf{0.3257±0.0016} & 0.3188±0.0012          & 0.3310±0.0006          & \textbf{0.3068±0.0012} & 0.3288±0.0015 & 0.3231±0.0007 & 0.3398±0.0008 \\ \midrule
\multirow{4}{*}{\rotatebox{90}{ILI}}        & 24  & 1.8344±0.0036          & \textbf{0.8229±0.0106} & \textbf{1.7857±0.0136} & 0.8289±0.0038          & 2.1226±0.1164          & 0.9201±0.0364 & 2.2966±0.0550 & 0.9567±0.0242 \\
                            & 36  & 1.7797±0.0111          & \textbf{0.8340±0.0069} & \textbf{1.7640±0.0550} & 0.8361±0.0187          & 1.8773±0.1362          & 0.9341±0.0619 & 2.2983±0.0507 & 0.9765±0.0319 \\
                            & 48  & 1.8154±0.0496          & 0.8614±0.0231          & \textbf{1.7011±0.1103} & \textbf{0.8445±0.0375} & 1.7393±0.0941          & 0.8852±0.0382 & 2.3440±0.0808 & 1.0248±0.0166 \\
                            & 60  & \textbf{1.7563±0.0528} & \textbf{0.8613±0.0187} & 1.9102±0.1089          & 0.9365±0.0389          & 1.8076±0.0763          & 0.9141±0.0220 & 2.1510±0.0124 & 0.9519±0.0083 \\ \bottomrule
\end{tabular}
\label{tab:error_bar}
}
\end{table*}
\section{Hyper-parameter Sensitivity}
\textbf{Influence of Individual factor} For Seasonal-TFB-IK in TFDNet-IK, we study the sensitivity of various individual factors $I$ on four datasets (Traffic, Electricity, Weather, and ETTm2) with a fixed prediction length $T=96$. The individual factor is set as $I=\{4,16,32,64\}$. As shown in Table \ref{tab:indivudual_factor_i},
different individual factors have no significant effects on the results, which verifies the model robustness with the individual factor $I$. 

%\resizebox{!}{\.5\paperheight}{
% \vskip -0.2in
\begin{table*}[h]
\centering
%\begin{footnotesize}
% \begin{adjustwidth}{-1.5in}{-1in}
\caption{TDFNet-IK performance under different choices of individual factor $I$ in the Seasonal-TFB-IK.}
\scalebox{0.9}{
\vspace{-1mm}
% \scalebox{0.60}{
\begin{tabular}{c|cccccccc}
\toprule
Individual Factor $I$ & \multicolumn{2}{c|}{4} & \multicolumn{2}{c|}{16} & \multicolumn{2}{c|}{32} & \multicolumn{2}{c}{64} \\ \midrule
Metric              & MSE        & MAE       & MSE        & MAE        & MSE        & MAE        & MSE        & MAE       \\ \midrule
Electricity         & 0.128      & 0.221     & 0.127      & 0.220      & 0.127      & 0.220      & 0.128      & 0.221     \\ \midrule
Traffic             & 0.374      & 0.252     & 0.370      & 0.250      & 0.374      & 0.250      & 0.377      & 0.256     \\ \midrule
Weather             & 0.143      & 0.189     & 0.142      & 0.188      & 0.143      & 0.188      & 0.143      & 0.188     \\ \midrule
ETTm2               & 0.157      & 0.244     & 0.157      & 0.243      & 0.157      & 0.244      & 0.157      & 0.244     \\ \bottomrule
\end{tabular}
\label{tab:indivudual_factor_i}
}
\end{table*}

\textbf{Influence of Kernel number} For Seasonal-TFB-MK in TFDNet-MK, we explore the effects of different kernel numbers on four datasets (Traffic, Electricity, Weather, and ETTm2). To trade-off performance and efﬁciency, we set $k=\{1,4,8,16\}$. As shown in Table \ref{tab:kernel_number_k}, models with multiple kernels have better performance than a single kernel. Especially for datasets with a large number of channels like Traffic, more kernels can improve the performance.

%\resizebox{!}{\.5\paperheight}{
% \vskip -0.2in
\begin{table*}[!tb]
\centering
%\begin{footnotesize}
% \begin{adjustwidth}{-1.5in}{-1in}
\caption{TDFNet-MK performance under different choices of kernel number $k$ in the Seasonal-TFB-MK.}
\scalebox{0.9}{
\vspace{-1mm}
% \scalebox{0.60}{
\begin{tabular}{c|cccccccc}
\toprule
Kernel number $k$ & \multicolumn{2}{c|}{1} & \multicolumn{2}{c|}{4} & \multicolumn{2}{c|}{8} & \multicolumn{2}{c}{16} \\ \midrule
Metric          & MSE        & MAE       & MSE        & MAE       & MSE        & MAE       & MSE        & MAE       \\ \midrule
Electricity     & 0.130      & 0.223     & 0.129      & 0.222     & 0.129      & 0.222     & 0.129      & 0.221     \\ \midrule
Traffic         & 0.368      & 0.250     & 0.362      & 0.248     & 0.361      & 0.247     & 0.354      & 0.241     \\ \midrule
Weather         & 0.151      & 0.204     & 0.148      & 0.194     & 0.150      & 0.204     & 0.153      & 0.206     \\ \midrule
ETTm2           & 0.158      & 0.245     & 0.158      & 0.246     & 0.158      & 0.245     & 0.160      & 0.247     \\ \bottomrule
\end{tabular}
\label{tab:kernel_number_k}
}
\end{table*}
\section{Additional Experiments}
%\subsection{Noise Injection Experiment}
%To demonstrate the robustness of our models in long-term time series forecasting tasks, we conduct the noise injection experiments following \cite{zhou2022film,wang2023micn} by adding a $0.3\mathcal{N}(0,1)$ Gaussian noise in the training and testing dataset. The results are shown in Table \ref{tab:noise}. The performance of TFDNet is not significantly impacted by the addition of Gaussian noise because the degradation is less than $3\%$ even in the worst case. However, we find injecting noise into the training dataset is able to improve our models' performance, which illustrates the robustness of our models. \ziyu{??}
%\input{Table/noise}

 \subsection{Loss Effect Analysis}
L2 loss function is commonly utilized in many time series forecasting models to measure the dissimilarity between the predicted and actual values \cite{wu2021autoformer,zhou2022fedformer,nie2022time}. We conduct experiments to study the effects of different loss functions. Our models are retrained by L2 loss function with the same settings and the results are in Table \ref{tab:loss}. However, L2 loss is sensitive to outliers which makes it is easier to affected by noises. Noise is abundant in real-time series, in this case, sensitivity to noise reduces the robustness of the model.

As shown in Table \ref{tab:loss}, our proposed Mixture loss can improve the prediction accuracy with lower MSE in most cases. And in all cases, Mixture loss function is able to make models more robust with lower MAE, compared with L2 loss.
%\resizebox{!}{\.5\paperheight}{
% \vskip -0.2in
\begin{table*}[t]
\centering
%\begin{footnotesize}
% \begin{adjustwidth}{-1.5in}{-1in}
\caption{Multivariate long-term series forecasting results from training with two different loss functions. ML represents our proposed Mixture loss function and L2 represents L2 loss function. \textbf{Bold} indicates the best.}
\scalebox{0.8}{
\vspace{-1mm}
% \scalebox{0.60}{
\begin{tabular}{cc|cccccccc}
\hline
\multicolumn{2}{c|}{Methods}                            & \multicolumn{2}{c}{TFDNet-IK(ML)} & \multicolumn{2}{c}{TFDNet-IK(L2)} & \multicolumn{2}{c}{TFDNet-MK(ML)} & \multicolumn{2}{c}{TFDNet-MK(L2)} \\ \hline
\multicolumn{2}{c|}{Metric}                             & MSE             & MAE             & MSE                  & MAE        & MSE             & MAE             & MSE                  & MAE        \\ \hline
\multicolumn{1}{c|}{\multirow{4}{*}{\rotatebox{90}{ETTh1}}}       & 96  & 0.359           & 0.386           & 0.364                & 0.394      & \textbf{0.356}  & \textbf{0.383}  & 0.361                & 0.390      \\
\multicolumn{1}{c|}{}                             & 192 & 0.398           & 0.412           & 0.395                & 0.416      & 0.396           & \textbf{0.409}  & \textbf{0.393}       & 0.411      \\
\multicolumn{1}{c|}{}                             & 336 & 0.432           & 0.431           & 0.424                & 0.437      & 0.431           & \textbf{0.429}  & \textbf{0.428}       & 0.434      \\
\multicolumn{1}{c|}{}                             & 720 & 0.438           & 0.453           & 0.455                & 0.475      & \textbf{0.421}  & \textbf{0.443}  & 0.439                & 0.464      \\ \hline
\multicolumn{1}{c|}{\multirow{4}{*}{\rotatebox{90}{ETTh2}}}       & 96  & 0.268           & 0.329           & 0.268                & 0.332      & 0.266           & \textbf{0.328}  & \textbf{0.265}       & 0.331      \\
\multicolumn{1}{c|}{}                             & 192 & 0.332           & 0.370           & 0.329                & 0.369      & 0.331           & \textbf{0.369}  & \textbf{0.329}       & 0.369      \\
\multicolumn{1}{c|}{}                             & 336 & 0.359           & \textbf{0.393}  & 0.354                & 0.393      & 0.361           & 0.394           & \textbf{0.353}       & 0.393      \\
\multicolumn{1}{c|}{}                             & 720 & 0.381           & 0.417           & 0.387                & 0.422      & \textbf{0.381}  & \textbf{0.416}  & 0.381                & 0.422      \\ \hline
\multicolumn{1}{c|}{\multirow{4}{*}{\rotatebox{90}{ETTm1}}}       & 96  & \textbf{0.283}  & \textbf{0.330}  & 0.283                & 0.339      & 0.286           & 0.333           & 0.288                & 0.343      \\
\multicolumn{1}{c|}{}                             & 192 & 0.326           & \textbf{0.355}  & \textbf{0.323}       & 0.365      & 0.327           & 0.356           & 0.326                & 0.365      \\
\multicolumn{1}{c|}{}                             & 336 & 0.359           & \textbf{0.377}  & \textbf{0.356}       & 0.385      & 0.360           & 0.379           & 0.359                & 0.387      \\
\multicolumn{1}{c|}{}                             & 720 & 0.412           & \textbf{0.408}  & \textbf{0.407}       & 0.414      & 0.410           & 0.408           & 0.409                & 0.414      \\ \hline
\multicolumn{1}{c|}{\multirow{4}{*}{\rotatebox{90}{ETTm2}}}       & 96  & \textbf{0.157}  & \textbf{0.244}  & 0.159                & 0.249      & 0.158           & 0.246           & 0.161                & 0.253      \\
\multicolumn{1}{c|}{}                             & 192 & 0.213           & \textbf{0.282}  & 0.212                & 0.287      & 0.214           & 0.283           & 0.215                & 0.290      \\
\multicolumn{1}{c|}{}                             & 336 & 0.264           & \textbf{0.318}  & 0.263                & 0.322      & 0.264           & 0.319           & 0.265                & 0.324      \\
\multicolumn{1}{c|}{}                             & 720 & \textbf{0.345}  & \textbf{0.371}  & 0.346                & 0.374      & 0.347           & 0.372           & 0.347                & 0.376      \\ \hline
\multicolumn{1}{c|}{\multirow{4}{*}{\rotatebox{90}{Electricity}}} & 96  & \textbf{0.128}  & \textbf{0.221}  & 0.129                & 0.226      & 0.129           & 0.221           & 0.129                & 0.225      \\
\multicolumn{1}{c|}{}                             & 192 & \textbf{0.145}  & \textbf{0.237}  & 0.145                & 0.240      & 0.147           & 0.239           & 0.147                & 0.242      \\
\multicolumn{1}{c|}{}                             & 336 & \textbf{0.160}  & \textbf{0.253}  & 0.161                & 0.257      & 0.163           & 0.256           & 0.164                & 0.260      \\
\multicolumn{1}{c|}{}                             & 720 & 0.197           & \textbf{0.285}  & \textbf{0.194}       & 0.290      & 0.200           & 0.288           & 0.200                & 0.292      \\ \hline
\multicolumn{1}{c|}{\multirow{4}{*}{\rotatebox{90}{Traffic}}}     & 96  & 0.377           & 0.256           & 0.379                & 0.275      & \textbf{0.354}  & \textbf{0.241}  & 0.361                & 0.261      \\
\multicolumn{1}{c|}{}                             & 192 & 0.392           & 0.263           & 0.395                & 0.281      & \textbf{0.372}  & \textbf{0.250}  & 0.379                & 0.269      \\
\multicolumn{1}{c|}{}                             & 336 & 0.406           & 0.266           & 0.409                & 0.286      & \textbf{0.388}  & \textbf{0.257}  & 0.393                & 0.275      \\
\multicolumn{1}{c|}{}                             & 720 & 0.447           & 0.287           & 0.444                & 0.306      & \textbf{0.428}  & \textbf{0.279}  & 0.433                & 0.295      \\ \hline
\multicolumn{1}{c|}{\multirow{4}{*}{\rotatebox{90}{Weather}}}     & 96  & \textbf{0.143}  & \textbf{0.188}  & 0.146                & 0.200      & 0.148           & 0.194           & 0.153                & 0.208      \\
\multicolumn{1}{c|}{}                             & 192 & \textbf{0.186}  & \textbf{0.230}  & 0.192                & 0.241      & 0.192           & 0.236           & 0.194                & 0.246      \\
\multicolumn{1}{c|}{}                             & 336 & \textbf{0.236}  & \textbf{0.269}  & 0.238                & 0.282      & 0.243           & 0.275           & 0.246                & 0.287      \\
\multicolumn{1}{c|}{}                             & 720 & 0.314           & \textbf{0.326}  & \textbf{0.313}       & 0.334      & 0.319           & 0.331           & 0.320                & 0.339      \\ \hline
\multicolumn{1}{c|}{\multirow{4}{*}{\rotatebox{90}{ILI}}}         & 24  & 1.834           & \textbf{0.823}  & 1.981                & 0.891      & \textbf{1.786}  & 0.829           & 1.824                & 0.860      \\
\multicolumn{1}{c|}{}                             & 36  & 1.780           & \textbf{0.834}  & 2.230                & 1.037      & \textbf{1.764}  & 0.836           & 1.784                & 0.860      \\
\multicolumn{1}{c|}{}                             & 48  & 1.815           & 0.861           & 2.188                & 1.028      & \textbf{1.701}  & \textbf{0.844}  & 1.938                & 0.956      \\
\multicolumn{1}{c|}{}                             & 60  & \textbf{1.756}  & \textbf{0.861}  & 2.045                & 0.983      & 1.910           & 0.936           & 2.176                & 1.037      \\ \hline
\end{tabular}
\label{tab:loss}
}
\end{table*}

\subsection{ Multi-scale Strategy Effect Analysis}
We employ the multi-scale strategy to adjust the sliding window size of STFT to capture the time-frequency information in diverse resolutions.
To analyze the effects of the multi-scale strategy with multiple STFT window lengths, we evaluate TFDNet with a single fixed window size on four datasets (Traffic, Electricity, Weather, ETTm2). As shown in Table \ref{muti_scale}, the two versions of TFDNet with multi-scale strategy have better performance on Traffic and Electricity datasets. However, it does not work well on ETTm2 dataset. 

\begin{figure}[!tb]
  \centering
  \begin{subfigure}[b]{0.32\textwidth}
    \includegraphics[width=\textwidth]{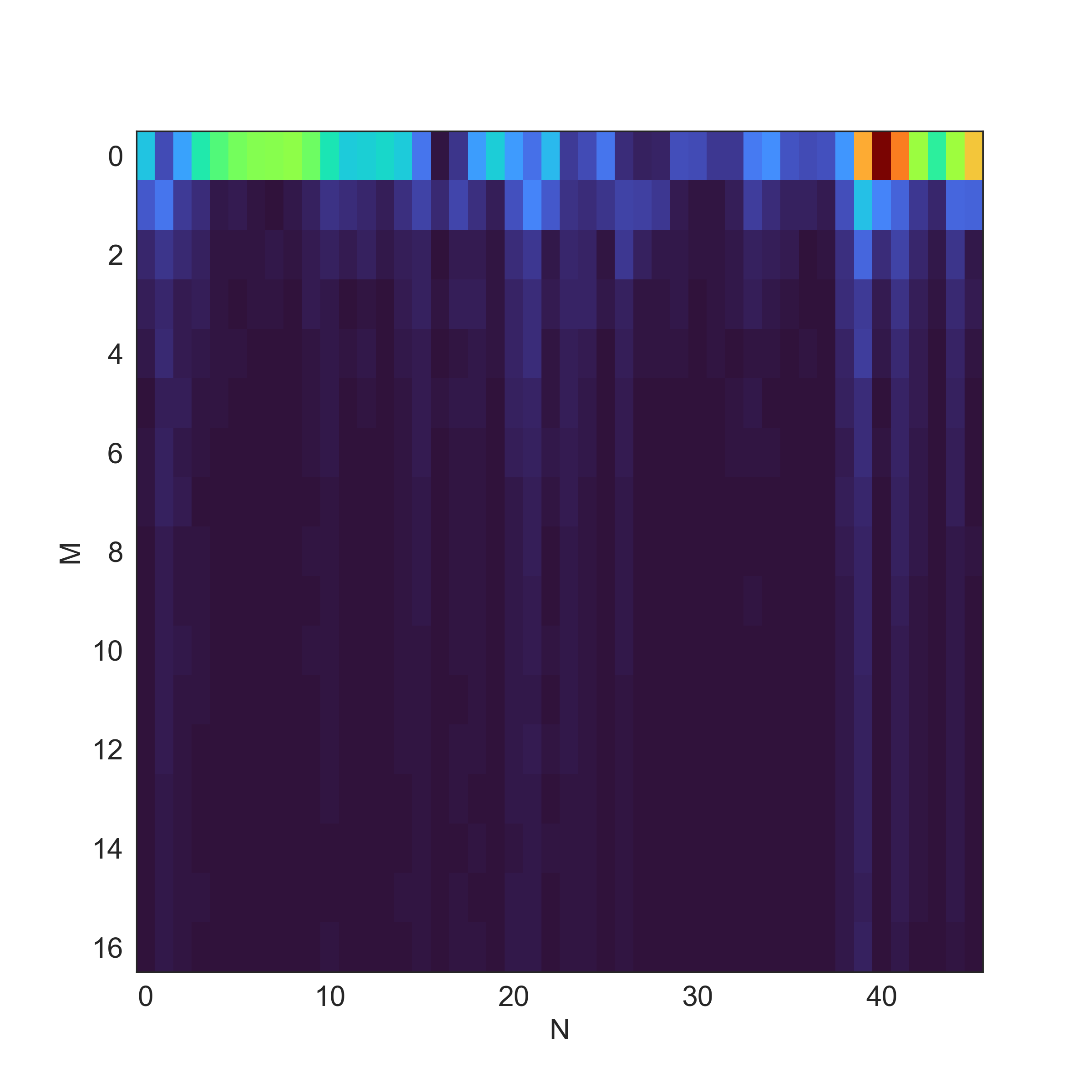}
    \caption{ETTm2 (Window length 32)}
  \end{subfigure}
  \begin{subfigure}[b]{0.32\textwidth}
    \includegraphics[width=\textwidth]{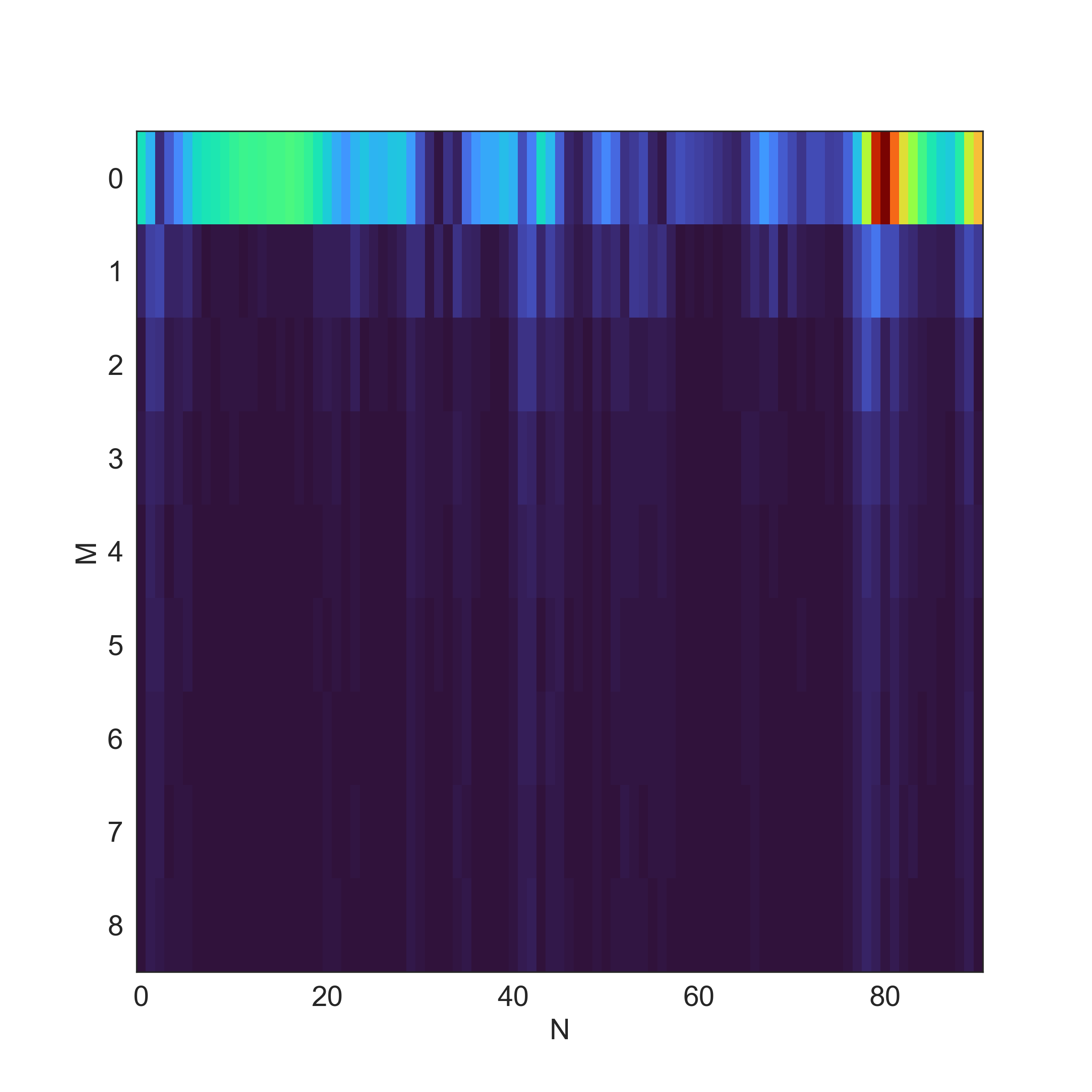}
    \caption{ETTm2 (Window length 16)}
  \end{subfigure}
  \begin{subfigure}[b]{0.32\textwidth}
    \includegraphics[width=\textwidth]{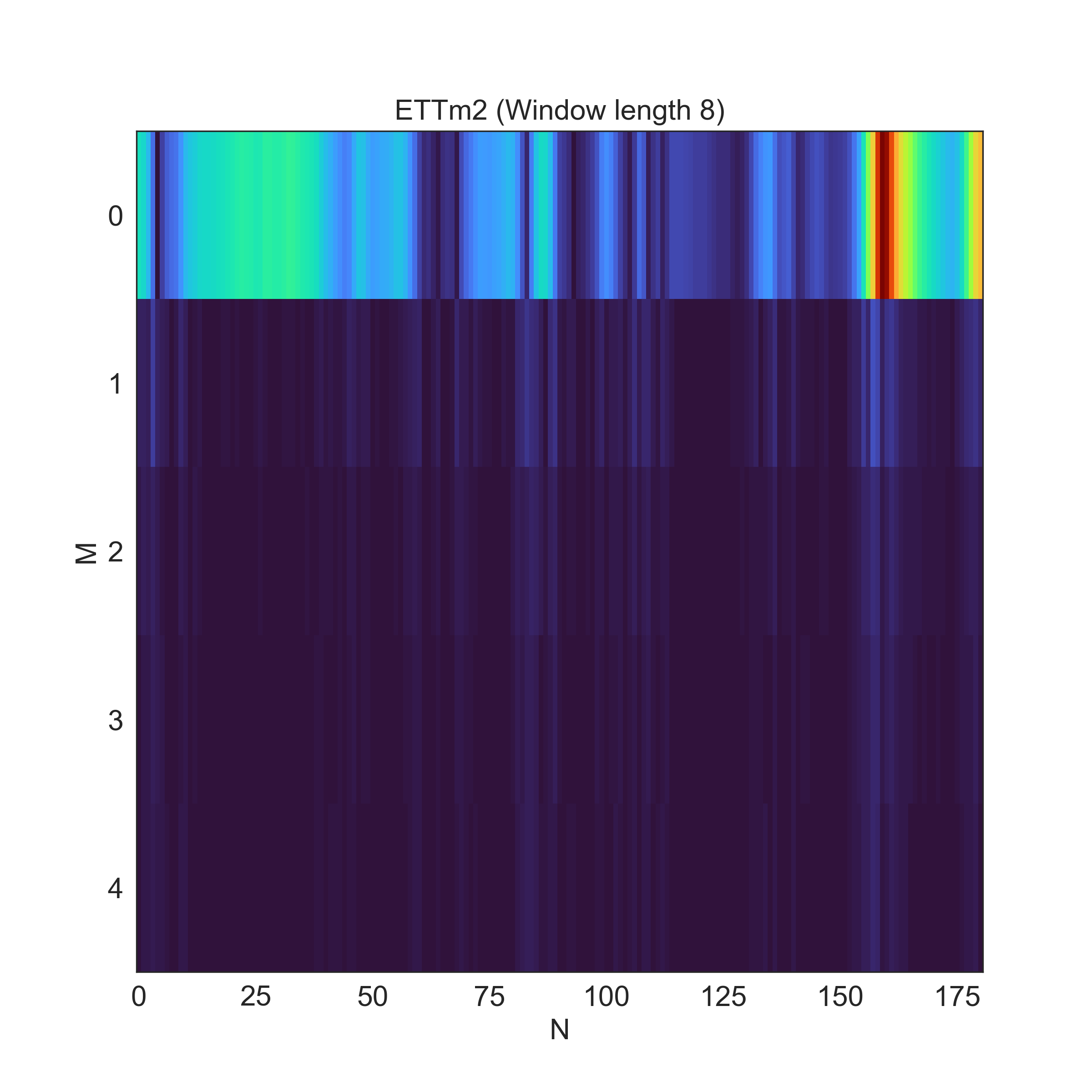}
    \caption{ETTm2 (Window length 8)}
  \end{subfigure}
  \begin{subfigure}[b]{0.32\textwidth}
    \includegraphics[width=\textwidth]{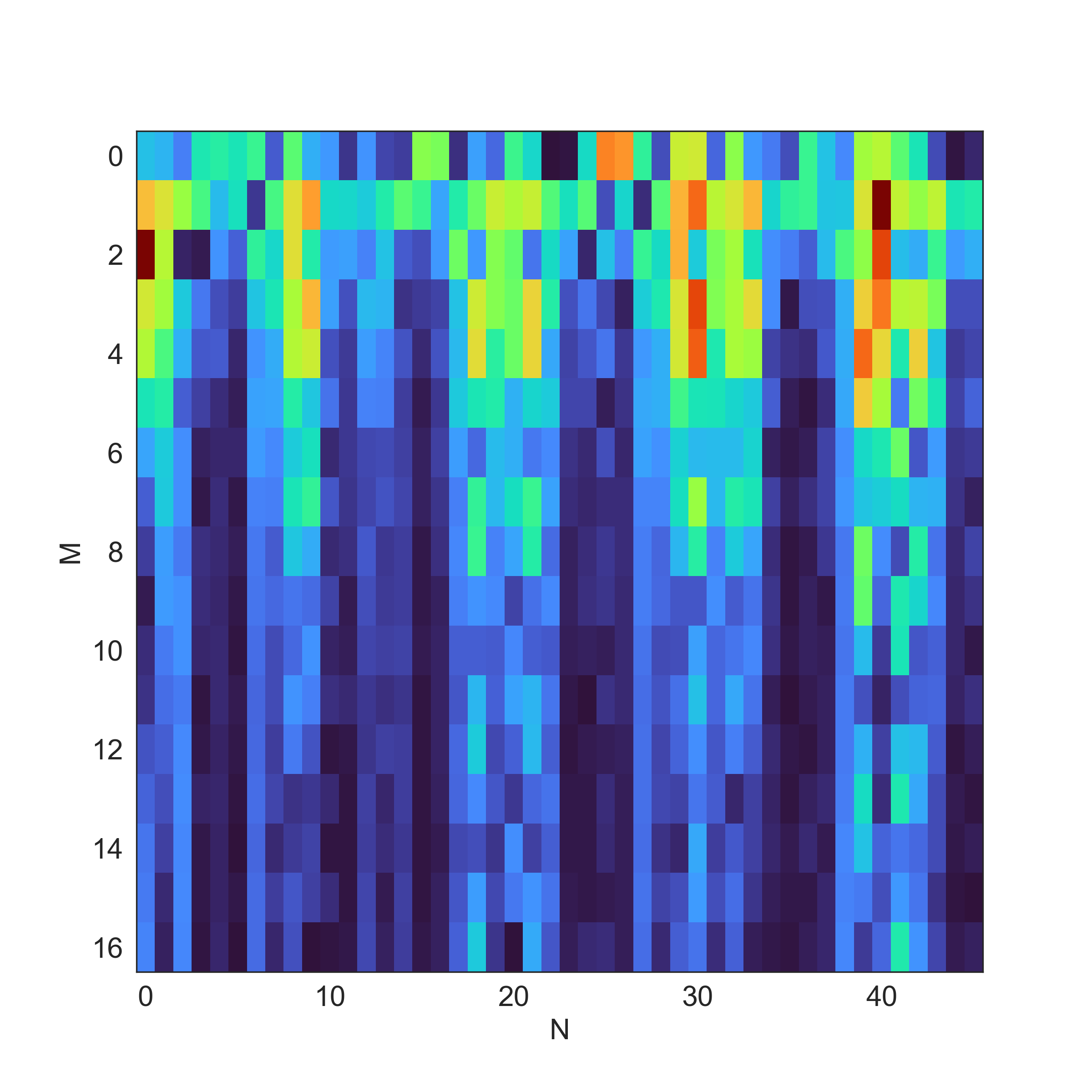}
    \caption{Traffic (Window length 32)}
  \end{subfigure}
  \begin{subfigure}[b]{0.32\textwidth}
    \includegraphics[width=\textwidth]{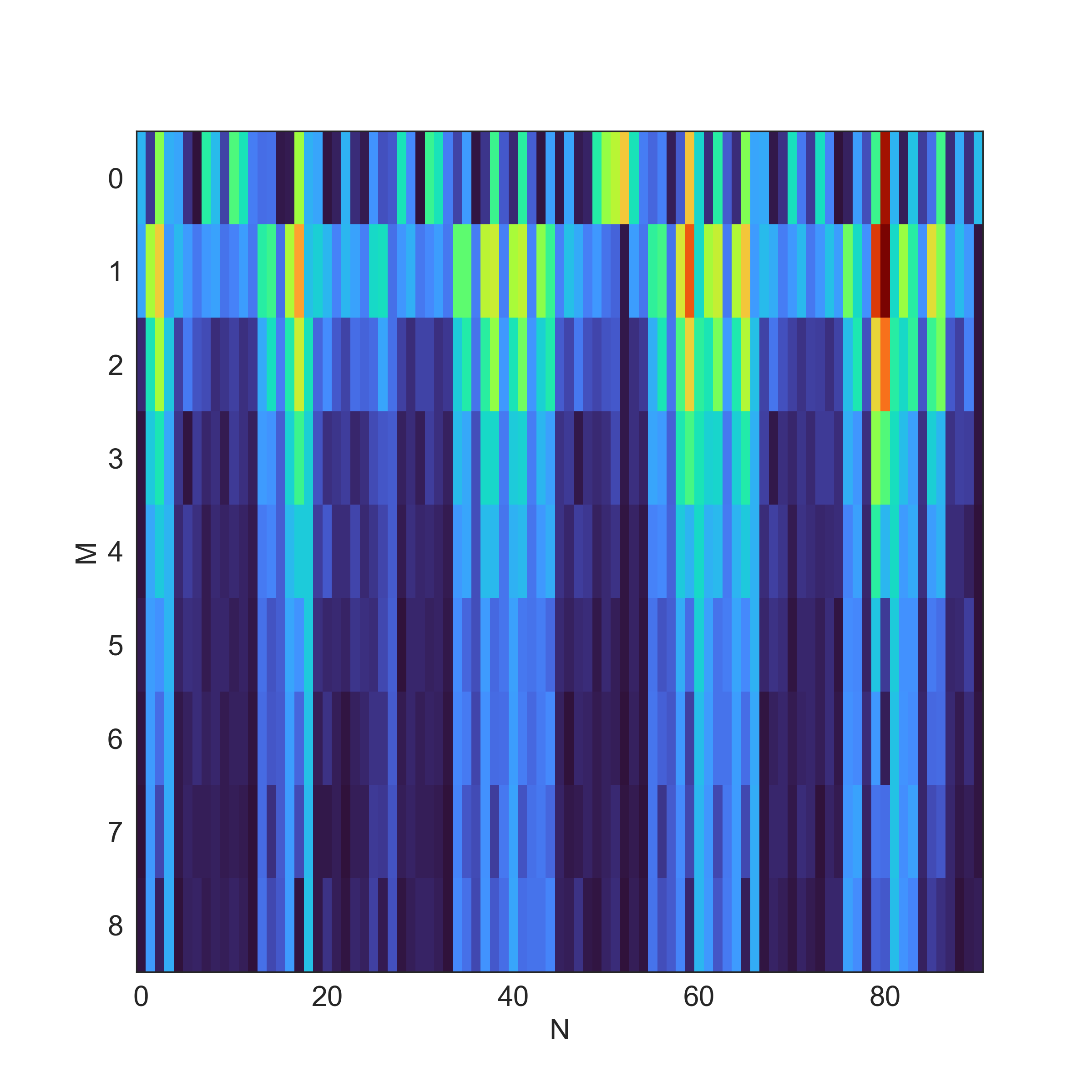}
    \caption{Traffic (Window length 16)}
  \end{subfigure}
  \begin{subfigure}[b]{0.32\textwidth}
    \includegraphics[width=\textwidth]{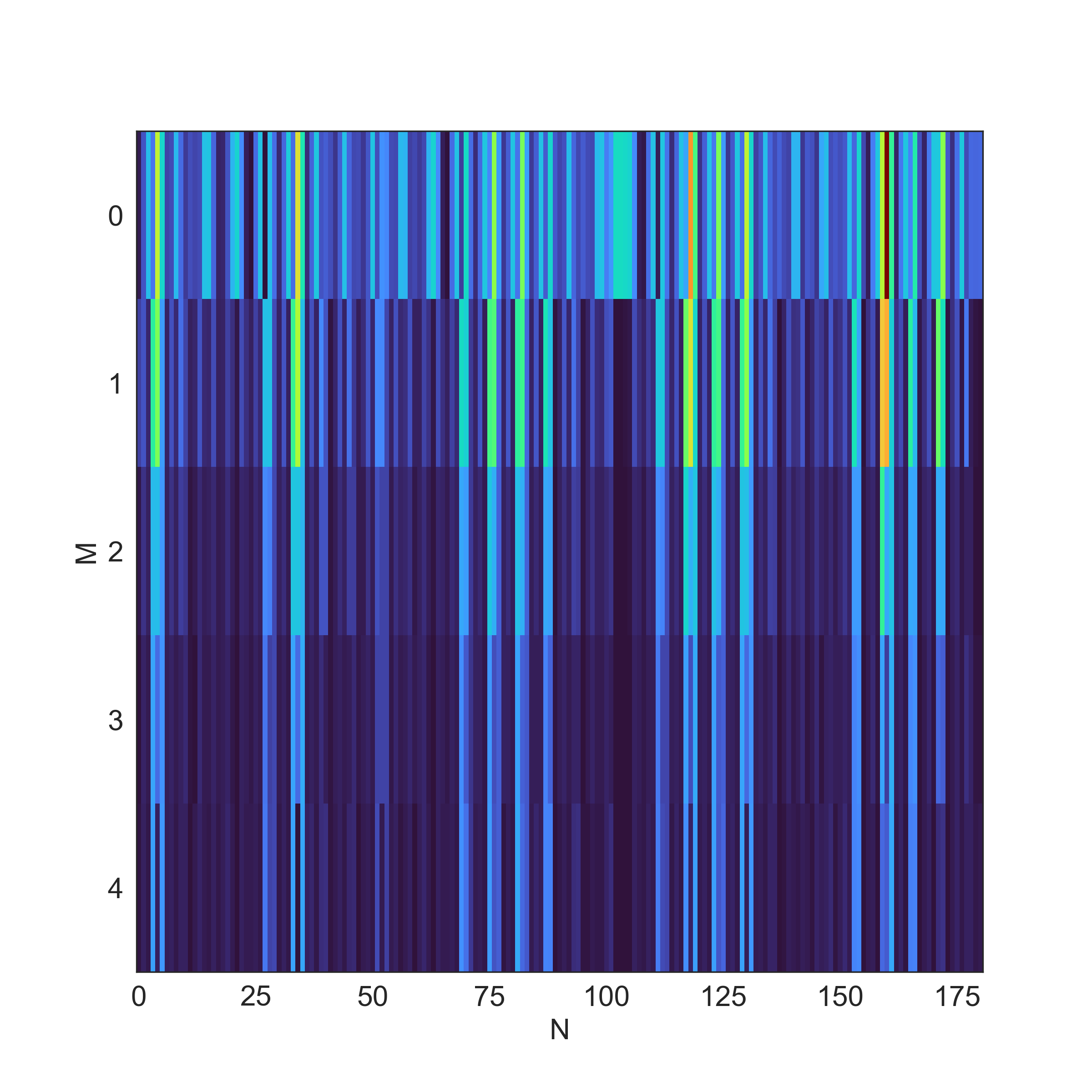}
    \caption{Traffic (Window length 8)}
  \end{subfigure}
  \caption{Time-frequency maps of ETTm2 and Traffic datasets with STFT window length $\{8,16,32\}$.}
  \label{ETTm2_tf_map}
\end{figure}

This phenomenon can be explained by their time-frequency maps. As shown in Figure \ref{ETTm2_tf_map},  the historical information of ETTm2 is mainly from the lowest frequency component in multi-resolutions,  with little information from the higher frequency components.  The historical information of Traffic dataset is distributed over different frequency components by comparison,  and the multi-scale windows are able to capture the time-frequency information in diverse resolutions.
% \usepackage{booktabs}
% \usepackage{multirow}
\begin{table*}[h]
\centering
\caption{The performance of TFDNet using the multi-scale strategy and the single strategy. Multi-scale strategy sets the STFT windows lengths $\windowsize$ as $\{8, 16, 32\}$, and single strategy sets $S$ as $16$. }
\vspace{-1mm}
\scalebox{0.58}{

\begin{tabular}{@{}cc|llll|llll|llll|llll@{}}
\toprule
\multicolumn{1}{l}{Datasets}                    & \multicolumn{1}{l|}{}                                                        & \multicolumn{4}{c|}{Traffic}                     & \multicolumn{4}{c|}{Electricity}                 & \multicolumn{4}{c|}{Weather}                     & \multicolumn{4}{c}{ETTm2}     \\ \midrule
\multicolumn{2}{l|}{Prediction Length T}                                            & 96    & 192   & 336   & \multicolumn{1}{c|}{720} & 96    & 192   & 336   & \multicolumn{1}{c|}{720} & 96    & 192   & 336   & \multicolumn{1}{c|}{720} & 96    & 192   & 336   & 720   \\ \midrule
\multicolumn{1}{c|}{\multirow{2}{*}{TFDNet-IK}}          & \multicolumn{1}{c|}{MSE} & 0.377 & 0.392 & 0.406 & 0.447                    & 0.128 & 0.145 & 0.160 & 0.197                    & 0.143 & 0.186 & 0.236 & 0.314                    & 0.157 & 0.213 & 0.264 & 0.345 \\
\multicolumn{1}{c|}{}                                    & \multicolumn{1}{c|}{MAE} & 0.256 & 0.263 & 0.266 & 0.287                    & 0.221 & 0.237 & 0.253 & 0.285                    & 0.188 & 0.230 & 0.269 & 0.326                    & 0.244 & 0.282 & 0.318 & 0.371 \\ \midrule
\multicolumn{1}{c|}{\multirow{2}{*}{TFDNet-IK (Single)}} & \multicolumn{1}{c|}{MSE} & 0.377 & 0.393 & 0.408 & 0.449                    & 0.128 & 0.146 & 0.162 & 0.200                    & 0.144 & 0.187 & 0.237 & 0.313                    & 0.157 & 0.211 & 0.264 & 0.346 \\
\multicolumn{1}{c|}{}                                    & \multicolumn{1}{c|}{MAE} & 0.256 & 0.264 & 0.270 & 0.290                    & 0.221 & 0.237 & 0.254 & 0.287                    & 0.189 & 0.230 & 0.270 & 0.325                    & 0.244 & 0.282 & 0.318 & 0.371 \\ \midrule
\multicolumn{1}{c|}{\multirow{2}{*}{TFDNet-MK}}          & \multicolumn{1}{c|}{MSE} & 0.354 & 0.372 & 0.388 & 0.428                    & 0.129 & 0.147 & 0.163 & 0.200                    & 0.148 & 0.192 & 0.243 & 0.319                    & 0.158 & 0.213 & 0.264 & 0.346 \\
\multicolumn{1}{c|}{}                                    & \multicolumn{1}{c|}{MAE} & 0.241 & 0.250 & 0.257 & 0.279                    & 0.221 & 0.239 & 0.256 & 0.288                    & 0.194 & 0.236 & 0.275 & 0.331                    & 0.246 & 0.284 & 0.319 & 0.372 \\ \midrule
\multicolumn{1}{c|}{\multirow{2}{*}{TFDNet-MK (Single)}} & \multicolumn{1}{c|}{MSE} & 0.361 & 0.379 & 0.394 & 0.433                    & 0.130 & 0.147 & 0.164 & 0.202                    & 0.148 & 0.193 & 0.243 & 0.317                    & 0.158 & 0.214 & 0.264 & 0.347 \\
\multicolumn{1}{c|}{}                                    & \multicolumn{1}{c|}{MAE} & 0.246 & 0.254 & 0.261 & 0.281                    & 0.222 & 0.239 & 0.256 & 0.289                    & 0.194 & 0.236 & 0.276 & 0.330                    & 0.246 & 0.283 & 0.319 & 0.372 \\ \bottomrule
\end{tabular}
}
\label{muti_scale}
\end{table*}

\subsection{Investigation of Trend-TFB}
Trend-TFB has a simpler structure and higher efficiency than Seaonal-TFB. To validate whether Trend-TFB is adequate to capture the trend information, we analyze the effects of the complex version of Trend-TFB with multiple kernels (similar to the multiple kernel strategy in Seasonal-TFB). Therefore, we have two additional versions, by changing the simple Trend-TFB with Trend-TFB with multiple kernels, respectively TFDNet-MK-IK and TFDNet-2MK for the subversions of Seasonal-TFB. TFDNet-MK-IK denotes to use Trend-TFB with multiple kernels and Seasonal-TFB-IK while TFDNet-2MK denotes to use Trend-TFB with multiple kernels and Seasonal-TFB-MK. The analysis results are shown in Table \ref{trend_tfb}. We can see that TFDNet-MK-IK using the multiple kernel strategy in the trend branch can not provide better performance compared with TFDNet-IK on datasets with low channel-wise correlations. Considering that TFDNet-IK is dedicated to forecasting time series with low correlations across channels, Trend-TFB with a simple structure is more applicable in terms of performance and efficiency. TFDNet-2MK also does not provide better performance than TFDNet-MK. The results verify that Trend-TFB with a single shared kernel is sufficient for processing trend information.
% \usepackage{booktabs}
% \usepackage{multirow}
\begin{table*}[t]
\centering
\caption{Investigation of Trend-TFB. TFDNet-MK-IK means using TFB with multiples kernels in TrendEncoder of TFDNet-IK, and TFDNet-2MK means using TFB with multiple kernels in both SeasonalEncoder and TrendEncoder }
\vspace{-1mm}
\scalebox{0.75}{
\begin{tabular}{@{}cc|cccccccccccc@{}}
\toprule
\multicolumn{1}{l}{Datasets}                       &     & \multicolumn{4}{c|}{Traffic}                     & \multicolumn{4}{c|}{Electricity}                 & \multicolumn{4}{c}{ETTm2}     \\ \midrule
\multicolumn{2}{l|}{Prediction Length T}                 & 96    & 192   & 336   & \multicolumn{1}{c|}{720} & 96    & 192   & 336   & \multicolumn{1}{c|}{720} & 96    & 192   & 336   & 720   \\ \midrule
\multicolumn{1}{c|}{\multirow{2}{*}{TFDNet-IK}}    & MSE & 0.377 & 0.392 & 0.406 & 0.447                    & 0.128 & 0.145 & 0.160 & 0.197                    & 0.157 & 0.213 & 0.264 & 0.345 \\
\multicolumn{1}{c|}{}                              & MAE & 0.256 & 0.263 & 0.266 & 0.287                    & 0.221 & 0.237 & 0.253 & 0.285                    & 0.244 & 0.282 & 0.318 & 0.371 \\ \midrule
\multicolumn{1}{c|}{\multirow{2}{*}{TFDNet-MK-IK}} & MSE & 0.373 & 0.388 & 0.404 & 0.445                    & 0.128 & 0.146 & 0.160 & 0.195                    & 0.157 & 0.216 & 0.267 & 0.347 \\
\multicolumn{1}{c|}{}                              & MAE & 0.255 & 0.261 & 0.268 & 0.287                    & 0.223 & 0.237 & 0.254 & 0.285                    & 0.244 & 0.283 & 0.320 & 0.372 \\ \midrule
\multicolumn{1}{c|}{\multirow{2}{*}{TFDNet-MK}}    & MSE & 0.354 & 0.372 & 0.388 & 0.428                    & 0.129 & 0.147 & 0.163 & 0.200                    & 0.158 & 0.214 & 0.264 & 0.347 \\
\multicolumn{1}{c|}{}                              & MAE & 0.241 & 0.250 & 0.257 & 0.279                    & 0.221 & 0.239 & 0.256 & 0.288                    & 0.246 & 0.283 & 0.319 & 0.372 \\ \midrule
\multicolumn{1}{c|}{\multirow{2}{*}{TFDNet-2MK}}   & MSE & 0.354 & 0.371 & 0.388 & 0.428                    & 0.129 & 0.147 & 0.164 & 0.199                    & 0.158 & 0.214 & 0.265 & 0.346 \\
\multicolumn{1}{c|}{}                              & MAE & 0.242 & 0.250 & 0.257 & 0.277                    & 0.221 & 0.238 & 0.257 & 0.287                    & 0.247 & 0.284 & 0.320 & 0.372 \\ \bottomrule
\end{tabular}
}
\label{trend_tfb}
\end{table*}
\section{Algorithm}
We present the details of algorithms designed in our paper. Algorithm~\ref{alg:TFDNet} demonstrates the overall process of TFDNet. Algorithm~\ref{alg:TE} and Algorithm~\ref{alg:SE} respectively describe the detailed process of TrendEncoder and SeasonalEncoder.
\begin{algorithm}[h]
  \setstretch{1.2}
  \caption{TFDNet}
\begin{algorithmic}[1]
    \STATE {\bfseries Input:} Input historical time series $\inputseq$; Input length $\inputlen$; Predict length $\predlen$; Data dimension $D$; STFT window $\windowsize$; Stride $\frameshift$. Technically, we set $\windowsize$ as $\{8,16,32\}$, $\frameshift$ as $\{3,8,16\}$.  
    % $R$ is randomly generated complex parameters, $\R\in\mathbb{C}^{N,d_{model},d_{model}}$.
  
\STATE $\inputseq=\text{RevIN}(\inputseq)^\mathsf{T}$\hfill{$ \inputseq\in \mathbb{R}^{D\times L}$}
\STATE $\inputseq_{tr}=\text{AvgPool}(\text{Paddding}(\inputseq))$\hfill{$ \inputseq_{tr}\in \mathbb{R}^{D\times L}$}
\STATE $\inputseq_{se}=\inputseq-\inputseq_{tr}$\hfill{$ \inputseq_{se}\in \mathbb{R}^{D\times L}$}
\STATE$\mathcal{\encoderout}_{se}=\text{Linear}(\text{SeasonEncoder}(\inputseq_{se}, S_1), \dots,\text{SeasonEncoder}(\inputseq_{se}, S_s))$\hfill{$ \encoderout_{se}\in \mathbb{R}^{D\times L}$}
\STATE$\mathcal{\encoderout}_{tr}=\text{Linear}(\text{TrendEncoder}(\inputseq_{tr}, S_1), \dots,\text{TrendEncoder}(\inputseq_{tr}, S_s))$\hfill{$ \encoderout_{tr}\in \mathbb{R}^{D\times L}$}
\STATE$\encoderout=\encoderout_{se}+\encoderout_{tr}$\hfill{$ \encoderout\in \mathbb{R}^{D\times L}$}
\STATE$\outputseq=\text{RevIN}(\text{Linear}(\encoderout)^\mathsf{T})$\hfill{$ \outputseq\in \mathbb{R}^{T\times D}$}
\STATE \textbf{Return} $\outputseq$
\end{algorithmic}
\label{alg:TFDNet}
\end{algorithm}

\begin{algorithm}[h]
\setstretch{1.2}
  \caption{TrendEncoder Procedure}
\begin{algorithmic}[1]
    \STATE{\bfseries Input:} Input trend component $\inputseq_{tr}$;Input length $\inputlen$;Data dimension $D$; STFT window $\windowsize_s$; Stride $\frameshift_s$.
    % $R$ is randomly generated complex parameters, $\R\in\mathbb{C}^{N,d_{model},d_{model}}$.
  
\STATE $\tilde{\inputseq}_{tr}=\text{STFT}(\inputseq_{tr},\windowsize_s,\frameshift_s)$\hfill{$ \tilde{\inputseq}_{tr}\in \mathbb{C}^{D\times M \times N}$}
\STATE \textbf{For} $i$ \textbf{in} $\{1,\cdots,D\}$: 
\STATE\quad $\TFBout_{tr}^{(i)}=\text{Kernel}(\STFTout^{(i)}_{tr}, \kernelmatrix_{tr})$ \hfill{$ \TFBout^{(i)}_{tr}\in \mathbb{C}^{ M \times N}$, $\kernelmatrix_{tr}\in \mathbb{C}^{M \times N \times N} $}
\STATE \textbf{End for} 
\STATE $\hat{\TFBout}_{tr} =\text{Linear}(\text{Reshape}(\TFBout_{tr}))$ \hfill{$ \hat{\TFBout}_{tr}\in \mathbb{C}^{D\times N \times M}$}
\STATE $\tilde{\encoderout}_{tr}=\TFBout_{tr}+\text{Reshape}(\text{Tanh}(\hat{\TFBout}_{tr})$ \hfill{$ \tilde{\encoderout}_{tr}\in \mathbb{C}^{D\times M \times N}$}
\STATE $\encoderout_{tr}=\text{STFT}^{-1}(\tilde{\encoderout}_{tr},\windowsize_s,\frameshift_s)$ \hfill{$ \encoderout\in \mathbb{R}^{D\times L}$}
\STATE \textbf{Return} $\encoderout$
\end{algorithmic}
\label{alg:TE}
\end{algorithm}

\begin{algorithm}[t]
\setstretch{1.2}
  \caption{SeasonalEncoder Procedure}
\begin{algorithmic}[1]
    \STATE{\bfseries Input:} Input seasonal component $\inputseq_{se}$;Input length $\inputlen$;Data dimension $D$; STFT window $\windowsize_s$; Stride $\frameshift_s$; Hyper-parameter $mode$; Hyper-parameter $I$; Hyper-parameter $k$.
    % $R$ is randomly generated complex parameters, $\R\in\mathbb{C}^{N,d_{model},d_{model}}$.
  
\STATE $\tilde{\inputseq}_{se}=\text{STFT}(\inputseq_ {se},\windowsize_s,\frameshift_s)$\hfill{$ \tilde{\inputseq}_{se}\in \mathbb{C}^{D\times M \times N}$}
\STATE \textbf{If} $mode==\text{IK}$:  
\STATE\quad $\kernelmatrix_{ind}=\kernelmatrix_1 \cdot \kernelmatrix_2$ \hfill{$\kernelmatrix_{ind} \in{\mathbb{\complex}^{\channels\times{\nfft}\times{\timeframe}\times{\timeframe}}}$, $\kernelmatrix_1\in{\mathbb{\complex}^{\channels\times{I}}}$,$\kernelmatrix_2\in{\mathbb{\complex}^{I\times{\nfft}\times{\timeframe}\times{\timeframe}}}$}
\STATE \quad\textbf{For} $i$ \textbf{in} $\{1,\cdots,D\}$: 
\STATE\quad\quad $\TFBout_{se}^{(i)}=\text{Kernel}(\STFTout_{se}^{(i)},\kernelmatrix_{ind}^{(i)})$ \hfill{$ \TFBout^{(i)}_ {se}\in \mathbb{C}^{ M \times N}$}
\STATE\quad \textbf{End for} 
\STATE\textbf{End if} 
\STATE \textbf{Else if} $mode==\text{MK}$:
\STATE \quad\textbf{for} $i$ \textbf{in} $\{1,\cdots,D\}$: 
\STATE\quad\quad \textbf{for} $j$ \textbf{in} $\{1,\cdots,k\}$: 
\STATE\quad\quad\quad$\hidden_j^{(i)}=\text{Kernel}(\STFTout^{(i)}_{se}, \kernelmatrix^j_{multi})$ \hfill{$ \hidden_j^{(i)}\in \mathbb{C}^{M \times N}$, $\kernelmatrix^j_{multi}\in \mathbb{C}^{M \times N \times N} $}
\STATE\quad\quad\quad$\gate_j=\text{Sigmod}(|\sum\gatematrix^j \odot \STFTout_{se}^{(i)}|)$ \hfill {$\gatematrix^j\in{\mathbb{\complex}^{\nfft\times{\timeframe}}}$}
\STATE\quad\quad \textbf{End for}
\STATE\quad\quad$\TFBout_{se}^{(i)}=\sum_k\gate \odot [\hidden_1^{(i)};\hidden_2^{(i)};\cdots \hidden_k^{(i)}]$\hfill{$ \TFBout^{(i)}_ {se}\in \mathbb{C}^{ M \times N}$}
\STATE\quad \textbf{End for} 
\STATE\textbf{End if} 
\STATE $\hat{\TFBout}_ {se} =\text{Linear}(\text{Reshape}(\TFBout_ {se}))$ \hfill{$ \hat{\TFBout}_ {se}\in \mathbb{C}^{D\times N \times M}$} 
\STATE $\tilde{\encoderout}_ {se}=\TFBout_ {se}+\text{Reshape}(\text{Tanh}(\hat{\TFBout}_ {se})$ \hfill{$ \tilde{\encoderout}_ {se}\in \mathbb{C}^{D\times M \times N}$}
\STATE $\encoderout_ {se}=\text{STFT}^{-1}(\tilde{\encoderout}_ {se},\windowsize_s,\frameshift_s)$ \hfill{$ \encoderout\in \mathbb{R}^{D\times L}$}
\STATE \textbf{Return} $\encoderout$
\end{algorithmic}
\label{alg:SE}
\end{algorithm}

% When having $k$ shared kernels, we learn $\{\kernelmatrix_{multi}^1,\cdots,\kernelmatrix_{multi}^k\}$ through the kernel operation for the multivariate time series. For the i-th channel time series, the output from one kernel is $\hidden_k^{(i)}=\text{Kernel}(\STFTout^{(i)}, \kernelmatrix^k_{multi})$, and we have multiple kernel representations $\{\hidden_1^{(i)}, \cdots,\hidden_k^{(i)}\}$. Next, the gate vector $\gate$ produced by $\gatematrix^k\in{\mathbb{\complex}^{\nfft\times{\timeframe}}}$ in the gate layer is utilized to fuse the multiple kernel representations. The $k$-th element in the gate vector is defined as
% $\gate_k=\text{Sigmod}(|\sum\gatematrix^k \odot \STFTout_{se}^{(i)}|)$.
% Finally, $\TFBout_{se}^{(i)}=\sum_k\gate \odot [\hidden_1^{(i)};\hidden_2^{(i)};\cdots \hidden_k^{(i)}]$ for  the $i$-th channel of the multivariate time series from Seasonal-TFB-MK.
\end{document}